\documentclass[final,5p,times,twocolumn]{elsarticle}
\biboptions{sort&compress}
\usepackage{amsmath,amssymb,amsfonts,mathrsfs}
\usepackage{graphicx,multirow,booktabs,array,tabularx,makecell}
\usepackage{xcolor,colortbl}
\usepackage{adjustbox}
\usepackage{placeins}
\usepackage{xurl}
\usepackage[hidelinks]{hyperref}
\journal{Image and Vision Computing}

\begin{document}
\begin{frontmatter}
\title{Cross-Encoder Anchor Refinement for Zero-Shot Anomaly Detection}
\author[hust]{Kaisong Ji\fnref{equal}}
\author[hust]{Ziyang Xu\fnref{equal}}
\author[nyu]{Chang Bian}
\author[hust]{Yu Zhou}
\author[hust]{Wenyu Liu}
\author[hust]{Xinggang Wang\corref{cor1}}
\ead{xgwang@hust.edu.cn}
\affiliation[hust]{organization={Huazhong University of Science and Technology},city={Wuhan},country={China}}
\affiliation[nyu]{organization={New York University},city={New York},country={USA}}
\fntext[equal]{These authors contributed equally to this work.}
\cortext[cor1]{Corresponding author.}

\begin{abstract}
Zero-shot anomaly detection aims to transfer anomaly discrimination from auxiliary training domains to unseen categories while localizing subtle defects. A key challenge is to construct discriminative references that combine contextual representations with complementary local evidence. We present Cross-Encoder Anchor Refinement (CEAR), which connects two visual encoding paths for the same image at the anchor stage, before normal and anomalous prototypes are constructed. An independent convolutional encoder pretrained for industrial product recognition builds an image-specific memory shared across selected vision transformer (ViT) layers. Spatial anchors read from ViT patch features query this memory through cross-attention and undergo residual refinement. Image-conditioned normal and anomalous queries then gate anchor aggregation to construct prototypes for anomaly scoring. Across 13 industrial and medical datasets, CEAR achieves the best performance in 57 of 68 evaluations spanning image-level detection and pixel-level localization. Notably, CEAR surpasses all other state-of-the-art methods in 41 of 44 pixel-level comparisons. Across 11 localization datasets, mean gains in pixel-level maximum F1 score and average precision are 4.45 and 3.79 percentage points, respectively. This study shows how cross-encoder interaction at the anchor stage can integrate complementary visual representations for fine-grained anomaly localization across domains.
\end{abstract}

\begin{keyword}
Zero-shot anomaly detection \sep Anomaly localization \sep Spatial anchor refinement \sep Image-conditioned prototypes \sep Cross-domain generalization
\end{keyword}
\end{frontmatter}

\section{Introduction}\label{sec1}

Industrial visual anomaly detection identifies defective products and localizes regions that deviate from normal appearance. Defects are rare and diverse, while changing product categories make extensive annotation and category-specific training costly. We study zero-shot anomaly detection (ZSAD) with auxiliary supervision, learning transferable anomaly discrimination for target categories absent from anomaly training. This setting couples cross-category recognition with precise localization, requiring sensitivity to fine cracks, small notches, and low-contrast regions within otherwise normal structures. MVTec AD, Real-IAD, and MVTec AD 2 illustrate these challenges through diverse defects and demanding imaging conditions~\citep{bergmann2019mvtec,wang2024real,heckler2026mvtec}.

Vision--language models provide transferable semantic references for ZSAD. WinCLIP combines prompt ensembles with window-, patch-, and image-level features, whereas AnomalyCLIP learns object-agnostic prompts for normality and abnormality~\citep{jeong2023winclip,zhou2024anomalyclip}. AA-CLIP, Bayes-PFL, and FAPrompt improve semantic separation, model prompt distributions, and adapt fine-grained abnormality prompts to individual images, respectively~\citep{ma2025aa,qu2025bayesian,zhu2025fine}. These advances also highlight the importance of matching semantic references to spatially faithful visual evidence. SCLIP and ClearCLIP improve dense CLIP predictions through attention reformulation and representation decomposition~\citep{wang2024sclip,lan2024clearclip}. For anomaly detection, FE-CLIP, DLVP-CLIP, and FB-CLIP exploit frequency cues, local visual prompts, and foreground--background separation~\citep{gong2025fe,zhang2026dlvp,hu2026fb}. Together, these studies motivate a closer examination of how local visual structure contributes to anomaly discrimination.

Recent work on pretrained visual representations highlights the importance of preserving spatial structure and recovering fine-grained detail for dense prediction~\citep{darcet2024vision,fu2024featup}. In anomaly detection, VisualAD constructs discriminative representations through visual tokens and spatial aggregation~\citep{hou2026visualad}. AnomalyVFM combines diverse anomaly synthesis with internal feature adaptation to improve transfer across datasets~\citep{fuvcka2026anomalyvfm}. These studies motivate our central question: how can contextual representations and complementary local evidence jointly inform transferable anomaly references?

\begin{figure}[!t]
\centering
\includegraphics[width=\columnwidth,height=0.79\textheight,keepaspectratio]{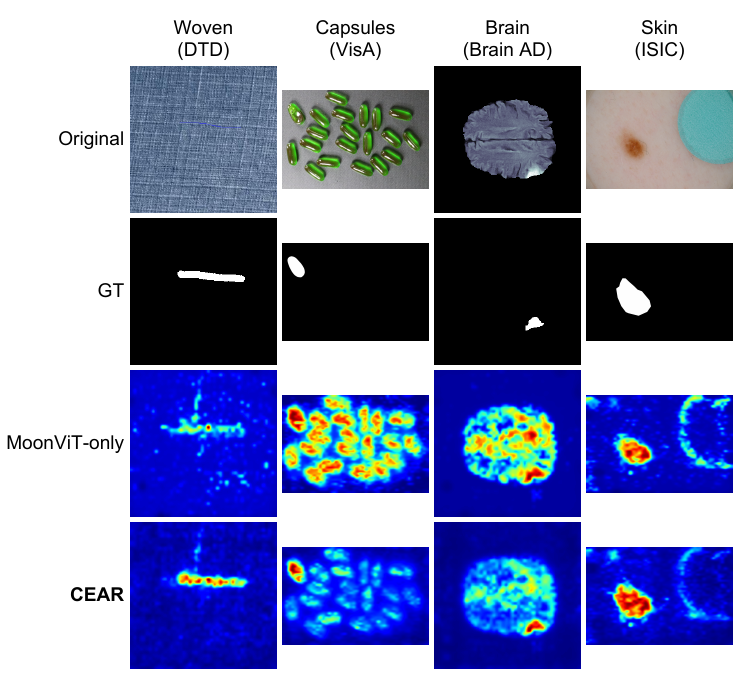}
\caption{Qualitative comparisons examine the contribution of cross-encoder conditioning to anomaly localization in industrial and medical images. CEAR concentrates responses more closely on annotated anomalies while reducing activations on some normal structures and background regions.}\label{fig1}
\end{figure}

Spatial sampling affects the local detail available for constructing these references. With a fixed patch size, higher input resolution produces a denser patch grid and can preserve evidence of small defects. We therefore use MoonViT, a visual encoder supporting variable input resolutions and pretrained on image--text pairs~\citep{team2025kimi}. Beyond spatial sampling, we focus on how local evidence can guide the construction of anomaly references at the output of this encoder. ViT-CoMer demonstrates that interactions between convolutional features and ViT representations can improve dense prediction~\citep{xia2024vit}. For ZSAD, we therefore examine how features from an independent local encoder can inform discriminative references derived from ViT representations. Spatial anchors provide the interaction interface: they retrieve local evidence and undergo refinement before query-gated aggregation constructs normal and anomalous prototypes. Section~\ref{sec4-3} compares this mechanism with score-level late fusion and direct feature concatenation under controlled conditions. Figure~\ref{fig1} compares anomaly responses from MoonViT-only and CEAR on industrial and medical samples, illustrating spatial differences associated with cross-encoder conditioning. In the examples shown, CEAR responds more selectively to annotated anomalies while reducing activations in some regions of normal texture and background.

We therefore present Cross-Encoder Anchor Refinement (CEAR), which connects two separate visual encoding paths through spatial anchors. Here, cross-encoder refers to refining anchors derived from ViT features using local evidence encoded independently from the same image. A ConvNeXt V2 CNN pretrained for industrial product recognition supplies hierarchical features~\citep{woo2023convnext}. These features form a local memory shared across selected MoonViT layers. Spatial anchors read from each selected layer query this memory through cross-attention and undergo residual refinement. The interaction operates on encoder outputs, before normal and anomalous prototypes are constructed. The anchors are shared by the normal and anomalous branches. Queries combining learned normal and anomalous seeds with image-level context then gate anchor aggregation to construct image-conditioned prototypes. The layerwise branches use independent refinement parameters and access the same memory; refined anchors are not passed between layers. Anomaly adaptation and inference use image features without text prompts or a language branch.

We evaluate CEAR on 13 industrial and medical datasets through cross-category and cross-dataset comparisons. Among the methods and protocols summarized in Table~\ref{tab1}, CEAR achieves 57 best results across 68 dataset--metric comparisons. For pixel-level localization, CEAR scores higher than all other state-of-the-art methods in 41 of 44 comparisons. Across 11 localization datasets, mean gains in pixel-level maximum F1 score (F1-max) and average precision (AP) are 4.45 and 3.79 percentage points, respectively. Mean gains in area under the receiver operating characteristic curve (AUROC) and area under the per-region overlap curve (PRO) are 2.20 and 2.39 percentage points. Beyond these performance comparisons, ablations vary industrial pretraining, memory features, interaction stage, and multilayer supervision. Component ablations and visualizations examine how local conditioning influences predictions.

Our main contributions are as follows.

\begin{itemize}

\item We introduce CEAR for zero-shot anomaly detection, integrating contextual representations and complementary local evidence through cross-encoder interaction before normal and anomalous prototypes are constructed.

\item We propose cross-encoder anchor refinement, allowing ViT-derived spatial anchors to retrieve evidence from a shared local memory built by an independent encoder. Layer-specific residual updates incorporate this evidence before query-gated construction of image-conditioned normal and anomalous prototypes.

\item We evaluate CEAR across 13 industrial and medical datasets. Matched-backbone and matched-resolution comparisons, component ablations, and qualitative analyses examine transfer performance and the role of local conditioning.

\end{itemize}

\section{Related Work}\label{sec2}

\subsection{Vision--Language References for Anomaly Detection}\label{subsec:related-vlm}

Vision--language methods use normal and anomalous text representations as transferable discriminative references. WinCLIP combines prompt ensembles with window-, patch-, and image-level matching, whereas AnomalyCLIP learns object-agnostic prompts~\citep{jeong2023winclip,zhou2024anomalyclip}. AdaCLIP combines static and image-conditioned prompts, while VCP-CLIP incorporates global and local visual context into text representations~\citep{cao2024adaclip,qu2024vcp}. FiLo couples fine-grained defect descriptions with dedicated localization modules~\citep{gu2024filo}. AA-CLIP strengthens anomaly-aware visual--textual representations, while AdaptCLIP alternately adapts the visual and textual branches~\citep{ma2025aa,gao2026adaptclip}.

Bayes-PFL, FAPrompt, and PromptMoE improve prompt flexibility through distribution modeling, sample-dependent abnormality prompts, and visually guided expert mixtures, respectively~\citep{qu2025bayesian,zhu2025fine,shao2026promptmoe}. FE-CLIP, DLVP-CLIP, and FB-CLIP enhance local discrimination through frequency-aware adaptation, regional visual prompting, and foreground--background separation, respectively~\citep{gong2025fe,zhang2026dlvp,hu2026fb}. These developments connect semantic adaptation with spatially localized evidence, motivating consideration of where such evidence enters anomaly reference construction.

\subsection{Reference Requirements and Transfer Settings}\label{subsec:related-visual}

Anomaly detectors differ in their access to target-domain information during training and inference. MuSc and MuSc-V2 exploit relationships among unlabeled test samples through mutual scoring~\citep{li2024musc,li2026musc}. RareCLIP estimates patch rarity using an online memory of previously observed images~\citep{he2025rareclip}. InCTRL, UniVAD, and MetaUAS use normal reference images at inference~\citep{zhu2024toward,gu2025univad,gao2024metauas}. Dinomaly and AnomalyMoE instead learn from target-category normal training data~\citep{guo2025dinomaly,gu2026anomalymoe}. Reference requirements and test-sample interactions are therefore distinct from whether a detector uses language.

For auxiliary-supervised transfer, AnomalyVFM combines anomaly synthesis with feature adaptation and confidence-weighted supervision~\citep{fuvcka2026anomalyvfm}. CEAR learns from auxiliary anomaly data and processes unseen target categories one image at a time. Its conditioning evidence comes from the current input, without target-domain normal references or interactions among test samples. Zero-shot evaluation here concerns target categories excluded from anomaly adaptation, while auxiliary training uses image labels and pixel masks.

\subsection{Dense Representations and Local Evidence}\label{subsec:related-dense}

Strong pretrained ViTs support dense prediction, as demonstrated by EoMT~\citep{kerssies2025your}, while MoonViT supports variable-resolution visual encoding~\citep{team2025kimi}. Register tokens and DVT address spatial artifacts in pretrained ViT features~\citep{darcet2024vision,yang2024denoising}. DINOv3 preserves dense feature quality through Gram anchoring~\citep{simeoni2025dinov3}. FeatUp and AnyUp recover spatial detail through feature upsampling~\citep{fu2024featup,wimmer2026anyup}. ViT-CoMer introduces multiscale CNN--ViT interactions within the backbone for dense prediction~\citep{xia2024vit}. These approaches improve spatial representations through different operations, from feature organization and upsampling to interactions between encoding paths.

Real-IAD and MVTec AD 2 highlight fine-scale defects and challenging industrial imaging conditions~\citep{wang2024real,heckler2026mvtec}. In CEAR, ConvNeXt V2 supplies hierarchical convolutional features~\citep{woo2023convnext}, with product-recognition pretraining on Real-IAD providing the industrial representation source. These features serve as local conditioning evidence at the output of the frozen ViT, before anomaly prototypes are constructed.

\subsection{Visual Prototypes and Conditional Refinement}\label{subsec:related-prototypes}

VisualAD constructs layerwise discriminative references through token-gated aggregation of spatial anchors, with attention and gates recomputed for each image~\citep{hou2026visualad}. UniADet learns task- and layer-specific decision weights for classification and segmentation~\citep{gao2026one}. ReCAP conditions prototypes on globally pooled features from the same visual encoder through bounded residual updates~\citep{wan2026beyond}. Layer-specific discrimination and image conditioning therefore provide established foundations for visual anomaly detection.

CEAR investigates how independently encoded local evidence can inform prototype construction. ViT-derived anchors query a shared CNN memory through cross-attention and undergo residual refinement before normal and anomalous queries gate their aggregation into prototypes. Relative to VisualAD and ReCAP, the distinction concerns both the independent source of local evidence and its application to anchors before prototype construction. The interaction operates on outputs of the frozen backbones; layerwise branches share the local memory but retain independent refinement parameters.

\section{Method}\label{sec3}

\subsection{Problem Formulation and Framework Overview}\label{sec3-1}

\begin{figure*}[!t]
\centering
\includegraphics[width=\textwidth,height=0.79\textheight,keepaspectratio]{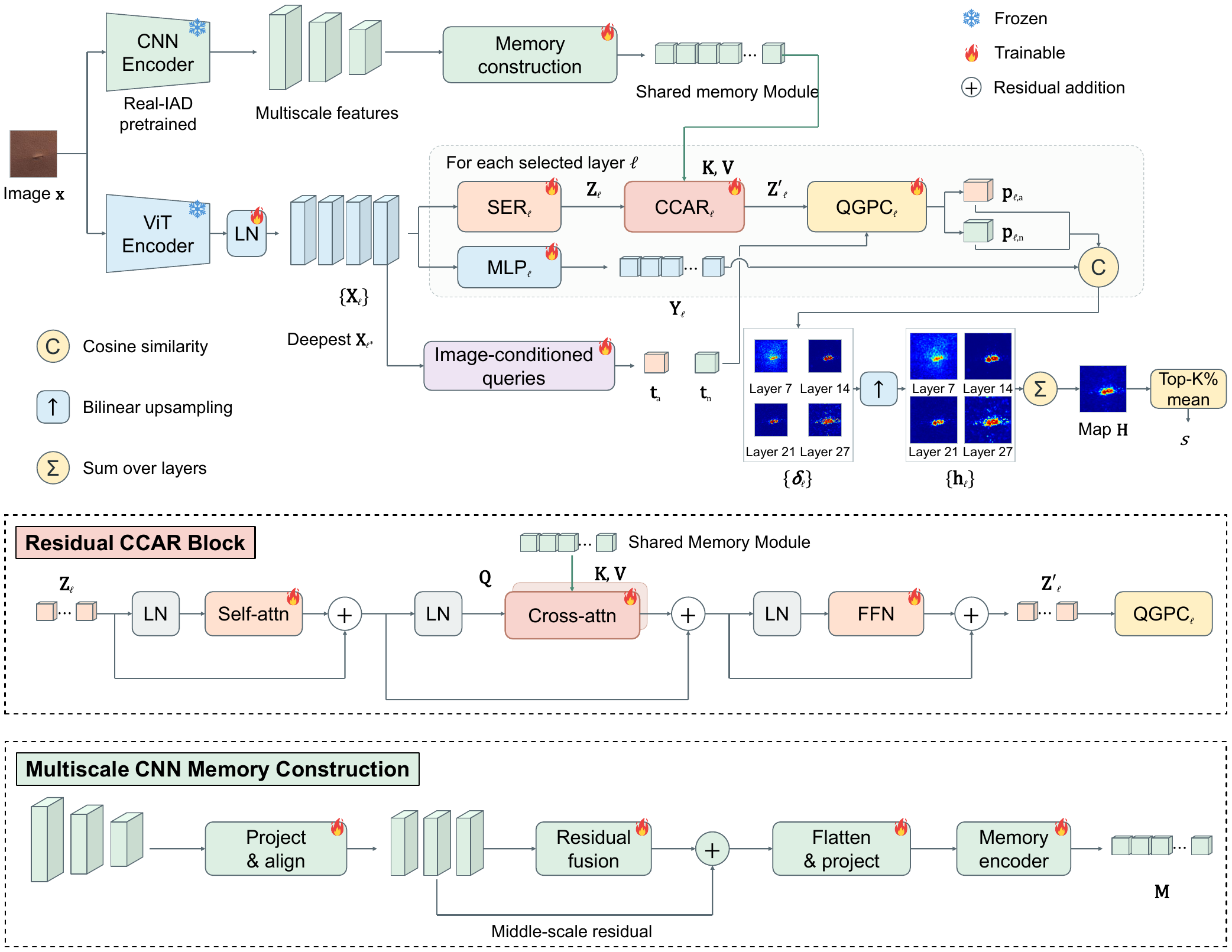}
\caption{CEAR architecture and cross-encoder interaction. \textbf{Top}: frozen ViT and CNN backbones encode the same image. The industrially pretrained CNN constructs a shared local memory for refining ViT-derived spatial anchors before query-gated construction of normal and anomalous prototypes. Projected patch features are compared with these prototypes to produce layerwise anomaly responses for localization and image-level scoring. \textbf{Middle}: the residual CCAR block. \textbf{Bottom}: multiscale CNN memory construction. Layerwise refinement shares the memory but uses independent parameters, without passing refined anchors between layers.}\label{fig2}
\end{figure*}

CEAR refines spatial anchors to connect ViT representations with local evidence encoded independently from the same image by a CNN. The ViT supplies patch features and contextual representations, while fused and encoded CNN features form a local memory shared across layerwise branches. At each selected ViT layer, spatial evidence readout (SER) first constructs anchors from patch features. CNN-conditioned anchor refinement (CCAR) then retrieves information from the shared memory to update these anchors. Finally, query-gated prototype construction (QGPC) aggregates the refined anchors into normal and anomalous prototypes. This cross-encoder interaction operates on anchors and memory derived from encoder outputs; refined anchors are not propagated between layers. Figure~\ref{fig2} illustrates the overall data flow, the residual structure of anchor refinement, and the construction of the shared local memory.

We consider an auxiliary training set \(\mathcal{D}_{s}=\{(\mathbf{x}_{i},y_{i},\mathbf{m}_{i})\}_{i=1}^{n_{s}}\), where image labels \(y_{i}\in \{0,1\}\) and pixel masks \(\mathbf{m}_{i}\in \{0,1\}^{H_{0}\times W_{0}}\) indicate anomaly status at their respective granularities. After learning from the auxiliary domain, the model directly predicts an image anomaly score \(s(\mathbf{x})\) and a pixel-level anomaly map \(\mathbf{H}(\mathbf{x})\) for unseen target categories. The CNN is pretrained using industrial product category supervision independent of both the source and target anomaly domains. Anomaly adaptation and inference operate entirely in the visual feature space, with each target image processed independently.

As shown in Figure~\ref{fig2}, the two encoders process the same image to obtain ViT patch features and multiscale CNN features. Global statistics from the deepest selected ViT layer generate image-conditioned queries. Within each layerwise branch, SER, CCAR, and QGPC successively read spatial evidence, refine anchors, and construct prototypes. A parallel patch projection MLP maps patch features into a space where cosine similarities to the two prototypes yield layerwise anomaly responses. These responses are bilinearly upsampled and summed into an anomaly map, whose highest-response pixels are averaged to obtain the image score. The frozen components in the figure denote the two encoder backbones. The trainable components comprise the shared output layer normalization and the anomaly modeling modules.

\subsection{Visual Representations and Shared Local Memory}\label{sec3-2}

Let \(\mathcal{I}\) denote the set of selected ViT layers. Shared layer normalization converts their outputs into patch features:

\begin{equation}\label{eq:cear-1}
\mathbf{X}_{\ell }=\operatorname{LN}_{o}(E_{\ell }(\mathbf{x}))\in \mathbb{R}^{N\times d},\ell \in \mathcal{I}.
\end{equation}

Here, \(E_{\ell }\) denotes the encoding mapping at layer \(\ell \), \(N\) is the number of patches, and \(d\) is the channel dimension. The selected layers retain the same spatial grid while providing representations at different depths. For clarity, we omit the batch dimension and training-time dropout. Tokens occupy rows of feature matrices, and linear transformations multiply weight matrices on the right.

To condition discrimination on the input image, we compute spatial means, maxima, and standard deviations from the deepest selected layer \(\mathbf{X}_{\ell ^{\ast }}\). A linear mapping converts these statistics into a global context vector \(\mathbf{u}\), which then conditions the normal and anomalous queries:

\begin{equation}\label{eq:cear-2}
\begin{aligned}
\mathbf{u}&=\bigl[\operatorname{mean}(\mathbf{X}_{\ell ^{\ast }});\max (\mathbf{X}_{\ell ^{\ast }}); \\
&\qquad\operatorname{std}(\mathbf{X}_{\ell ^{\ast }})\bigr]\mathbf{W}_{u}+\mathbf{b}_{u}, \\
\mathbf{t}_{c}&=\mathbf{q}_{c}+\mathbf{u}\mathbf{W}_{c}+\mathbf{b}_{c},c\in \{a,n\}.
\end{aligned}
\end{equation}

Here, \([\cdot ;\cdot ;\cdot ]\) denotes channel concatenation, \(\ell ^{\ast }=\max \mathcal{I}\) identifies the deepest selected layer, and \(c\in \{a,n\}\) indexes the anomalous and normal classes, respectively. Each learnable query seed \(\mathbf{q}_{c}\) is combined with image context to form the corresponding image-conditioned query \(\mathbf{t}_{c}\). The three statistics capture average responses, salient activations, and spatial response variability. They allow the queries to combine a discriminative direction shared across images with information specific to the current input. These queries are generated at the output of the encoder.

The CNN branch in Figure~\ref{fig2} extracts multiscale features \(\{\mathbf{C}_{s}\}_{s\in \mathcal{S}}\) representing texture variations, edges, and component structure in the input image. Channel projection \(\mathcal{P}_{s}\) and spatial alignment \(\mathcal{A}_{s}\) map the features onto a common grid. Residual fusion then uses the intermediate scale \(b\) as its reference:

\begin{equation}\label{eq:cear-3}
\begin{aligned}
\mathbf{F}_{s}&=\mathcal{A}_{s}(\mathcal{P}_{s}(\mathbf{C}_{s})),s\in \mathcal{S}, \\
\mathbf{F}&=\mathbf{F}_{b}+\beta _{f}\Phi _{f}(\operatorname{Concat}_{s\in \mathcal{S}}(\mathbf{F}_{s})).
\end{aligned}
\end{equation}

All aligned features \(\mathbf{F}_{s}\) have shape \(h_{m}\times w_{m}\times d_{c}\). The convolutional fusion mapping \(\Phi _{f}\) supplies complementary features, whose contribution is controlled by \(\beta _{f}\). The residual path preserves the intermediate-scale representation, while the fusion mapping learns complementary information from the other scales.

The ``Flatten \& project'' block in Figure~\ref{fig2} flattens the fused features and matches their channel dimension to that of the ViT. The ``Memory encoder'' then uses self-attention to model relationships among local structures, producing the shared memory:

\begin{equation}\label{eq:cear-4}
\begin{aligned}
\mathbf{T}&=\operatorname{Flatten}(\mathbf{F})\mathbf{W}_{m}, \\
\mathbf{M}&=\operatorname{LN}_{2}(\mathbf{T}+\operatorname{MHSA}(\operatorname{LN}_{1}(\mathbf{T}))).
\end{aligned}
\end{equation}

The resulting memory has dimensions \(\mathbf{M}\in \mathbb{R}^{N_{m}\times d}\), with \(N_{m}=h_{m}w_{m}\), and MHSA denotes multihead self-attention. Computed from the current image, this memory is shared across all selected ViT layers. Each layer retrieves the information it needs through an independent conditional refinement module.

\subsection{Cross-Encoder Anchor Refinement and Discriminative Prototypes}\label{sec3-3}

SER compresses the patch sequence into a set of spatial anchors. Each layer has learnable anchor seeds \(\mathbf{A}_{\ell }\in \mathbb{R}^{K\times d}\), from which queries, keys, and values are constructed for spatial readout:

\begin{equation}\label{eq:cear-5}
\begin{aligned}
\mathbf{Q}_{\ell }&=\operatorname{LN}_{q}(\mathbf{A}_{\ell }\mathbf{W}_{q,\ell }), \\
\mathbf{K}_{\ell }&=\operatorname{LN}_{k}((\mathbf{X}_{\ell }+\mathbf{P}_{\ell })\mathbf{W}_{k,\ell }), \\
\mathbf{V}_{\ell }&=\mathbf{X}_{\ell }\mathbf{W}_{v,\ell }.
\end{aligned}
\end{equation}

The positional embedding \(\mathbf{P}_{\ell }\in \mathbb{R}^{N\times d}\) is added only to the key branch. Queries and keys are layer-normalized after projection. Spatial readout weights then aggregate the projected patch features into anchors:

\begin{equation}\label{eq:cear-6}
\begin{aligned}
\mathbf{B}_{\ell }&=\operatorname{softmax}(\mathbf{Q}_{\ell }\mathbf{K}_{\ell }^{\top }/\sqrt{d}), \\
\mathbf{Z}_{\ell }&=\mathbf{B}_{\ell }\mathbf{V}_{\ell }.
\end{aligned}
\end{equation}

The softmax operation normalizes over patches, yielding spatial readout weights \(\mathbf{B}_{\ell }\in \mathbb{R}^{K\times N}\) and aggregated anchors \(\mathbf{Z}_{\ell }\in \mathbb{R}^{K\times d}\). These anchors provide intermediate evidence shared by the normal and anomalous branches. Class-specific discriminative prototypes are constructed subsequently through query-gated aggregation.

CCAR takes \(\mathbf{Z}_{\ell }\) as input and sequentially applies anchor self-attention, cross-attention to the CNN memory, and a feed-forward update (Figure~\ref{fig2}):

\begin{equation}\label{eq:cear-7}
\begin{aligned}
\mathbf{Z}_{\ell }^{(1)}&=\mathbf{Z}_{\ell }+\operatorname{MHSA}_{\ell }(\operatorname{LN}_{s,\ell }(\mathbf{Z}_{\ell })), \\
\mathbf{Z}_{\ell }^{(2)}&=\mathbf{Z}_{\ell }^{(1)}+\operatorname{MHCA}_{\ell }(\operatorname{LN}_{q,\ell }(\mathbf{Z}_{\ell }^{(1)}),\mathbf{M},\mathbf{M}), \\
\mathbf{Z}_{\ell }^{\prime }&=\mathbf{Z}_{\ell }^{(2)}+\operatorname{FFN}_{\ell }(\operatorname{LN}_{f,\ell }(\mathbf{Z}_{\ell }^{(2)})).
\end{aligned}
\end{equation}

MHCA denotes multihead cross-attention~\citep{vaswani2017attention}, with normalized anchors as queries and the shared memory \(\mathbf{M}\) as keys and values. Self-attention models relationships among anchors, whereas cross-attention retrieves local features relevant to the current ViT evidence from the independently encoded memory. The feed-forward network then applies a nonlinear transformation to the updated anchors. All three sublayers use residual connections, and each selected layer has independent refinement parameters. The refined anchors \(\mathbf{Z}_{\ell }^{\prime }\) combine ViT evidence with CNN conditioning information and directly support prototype construction.

QGPC uses the normal and anomalous queries to select and aggregate evidence from the refined anchors:

\begin{equation}\label{eq:cear-8}
\begin{aligned}
\mathbf{g}_{\ell ,c}&=\sigma (\mathbf{t}_{c}\mathbf{W}_{g,\ell }+\mathbf{b}_{g,\ell }), \\
\mathbf{p}_{\ell ,c}&=\mathbf{t}_{c}+\alpha _{\ell }((\mathbf{g}_{\ell ,c}\mathbf{Z}_{\ell }^{\prime })\mathbf{W}_{o,\ell }+\mathbf{b}_{o,\ell }).
\end{aligned}
\end{equation}

Here, \(\mathbf{g}_{\ell ,c}\in (0,1)^{1\times K}\) contains independent sigmoid gates, and \(\alpha _{\ell }\) is a learnable residual coefficient. The matrix product \(\mathbf{g}_{\ell ,c}\mathbf{Z}_{\ell }^{\prime }\) weights and aggregates the refined anchors. Together with the two queries, this aggregation yields image-conditioned normal and anomalous prototypes \(\mathbf{p}_{\ell ,c}\) at each layer. The two class branches share parameters for spatial readout and conditional refinement; their distinction is expressed through the queries and gating weights.

CCAR introduces local evidence from the independent CNN into the ViT-derived anchors. QGPC then combines image-conditioned queries with gated anchor contributions to construct normal and anomalous prototypes. The two encoding paths therefore interact through intermediate anchors before prototype aggregation. The CNN serves as a source of local evidence in this design. Section~\ref{sec4-3-1} compares the CNN-only model, fusion controls, and full model to examine how standalone detection performance relates to its benefit as a conditioning source.

\subsection{Patch Comparison and Anomaly Scoring}\label{sec3-4}

A patch projection MLP maps each layer's patch features into a space compatible with its discriminative prototypes. It contains two linear layers that preserve the channel dimension, separated by a GELU activation. For the projected features \(\mathbf{Y}_{\ell }=\operatorname{MLP}_{\ell }(\mathbf{X}_{\ell })\), the difference in cosine similarity to the anomalous and normal prototypes defines the patch anomaly response:

\begin{equation}\label{eq:cear-9}
\delta _{\ell ,i}=\cos (\mathbf{Y}_{\ell ,i},\mathbf{p}_{\ell ,a})-\cos (\mathbf{Y}_{\ell ,i},\mathbf{p}_{\ell ,n}).
\end{equation}

Here, \(i\) indexes patch positions, and \(\cos (\cdot ,\cdot )\) denotes cosine similarity. The response \(\delta _{\ell ,i}\) increases as a patch becomes more similar to the anomalous prototype or less similar to the normal prototype.

We reshape each layer's responses into a spatial grid and bilinearly upsample them. An unweighted sum forms the pixel-level anomaly map, and the image score averages a fraction of its highest-response pixels:

\begin{equation}\label{eq:cear-10}
\mathbf{H}(\mathbf{x})=\sum _{\ell \in \mathcal{I}} \mathcal{U}(\operatorname{Reshape}(\delta _{\ell })),
\end{equation}

\begin{equation}\label{eq:cear-11}
s(\mathbf{x})=\frac{1}{k}\sum _{i\in \mathcal{T}_{k}(\mathbf{H}(\mathbf{x}))} H_{i}(\mathbf{x}).
\end{equation}

Here, \(\mathcal{U}\) upsamples the patch grid to \(H_{0}\times W_{0}\), and \(\mathcal{T}_{k}(\mathbf{H})\) returns the \(k\) highest-response locations, where \(k=\max (1,\lfloor \rho H_{0}W_{0}\rfloor )\). The fraction \(\rho \) determines how many pixels contribute to image-level discrimination. This aggregation retains the contribution of localized anomalies while reducing sensitivity to a single extreme response.

\subsection{Joint Training Objective}\label{sec3-5}

The training objective combines task prediction, discriminative margins, and region completeness. We use binary cross-entropy for image-level prediction and supervise the deeper branches \(\mathcal{I}_{s}\subset \mathcal{I}\) with focal and Dice segmentation losses~\citep{lin2017focal,milletari2016v}:

\begin{equation}\label{eq:cear-12}
\mathcal{L}_{\mathrm{cls}}=\operatorname{BCEWithLogits}(s(\mathbf{x}),y),
\end{equation}

\begin{equation}\label{eq:cear-13}
\begin{aligned}
\mathcal{L}_{\mathrm{seg}}&=\sum _{\ell \in \mathcal{I}_{s}} \bigl[\mathcal{L}_{\mathrm{focal}}(\sigma (\mathbf{h}_{\ell }),\mathbf{m}) \\
&\qquad+\mathcal{L}_{\mathrm{Dice}}(\sigma (\mathbf{h}_{\ell }),\mathbf{m})\bigr].
\end{aligned}
\end{equation}

Here, \(\mathbf{h}_{\ell }=\mathcal{U}(\operatorname{Reshape}(\delta _{\ell }))\), \(\sigma \) is the sigmoid function, and \(\mathbf{m}\) is the ground-truth pixel mask. Segmentation losses are computed separately for each layer and summed over the selected deeper branches. A directional margin on the image-conditioned queries maintains separation between the normal and anomalous classes:

\begin{equation}\label{eq:cear-14}
\mathcal{L}_{\mathrm{qry}}=\mathbb{E}_{\mathbf{x}}[\max (0,\cos (\mathbf{t}_{a},\mathbf{t}_{n})+\tau )].
\end{equation}

We use the positive-part notation \([z]_{+}=\max (0,z)\), and \(\tau \) specifies the query margin. Applied to the queries \(\mathbf{t}_{a},\mathbf{t}_{n}\), this loss penalizes similarity above the prescribed boundary.

We additionally impose region-level margin supervision on the patch grid. Nearest-neighbor downsampling of the ground-truth mask defines anomalous locations \(\Omega ^{+}\) and normal locations \(\Omega ^{-}\). For each image, the region loss is defined as

\begin{equation}\label{eq:cear-15}
\begin{aligned}
&\mathcal{R}_{\ell }(\eta ^{+},\eta ^{-},w,r) \\
&\quad=\mathcal{H}_{r}(\{[\eta ^{+}-\delta _{\ell ,i}]_{+}:i\in \Omega ^{+}\}) \\
&\qquad+w\mathcal{H}_{r}(\{[\eta ^{-}+\delta _{\ell ,i}]_{+}:i\in \Omega ^{-}\}).
\end{aligned}
\end{equation}

The margins \(\eta ^{+}\) and \(\eta ^{-}\) correspond to foreground and background, respectively, and \(w\) weights the background term. The operator \(\mathcal{H}_{r}\) averages the largest fraction \(r\) of values in its input set. Hard examples are selected independently for foreground and background, and each term is averaged over images containing a nonempty corresponding region. Images with an empty region are excluded from that term's reduction. Different parameters are used for the deeper branches and the shallower branch set \(\mathcal{I}_{t}\):

\begin{equation}\label{eq:cear-16}
\begin{aligned}
\mathcal{L}_{\mathrm{sem}}&=\frac{1}{|\mathcal{I}_{s}|}\sum _{\ell \in \mathcal{I}_{s}} \mathcal{R}_{\ell }^{\mathrm{batch}}(\eta _{s}^{+},\eta _{s}^{-},w_{s},r_{s}), \\
\mathcal{L}_{\mathrm{tex}}&=\frac{1}{|\mathcal{I}_{t}|}\sum _{\ell \in \mathcal{I}_{t}} \mathcal{R}_{\ell }^{\mathrm{batch}}(\eta _{t}^{+},\eta _{t}^{-},w_{t},r_{t}).
\end{aligned}
\end{equation}

Here, \(\mathcal{R}_{\ell }^{\mathrm{batch}}\) denotes the region loss after the image-wise reduction described above. The deeper-layer constraint promotes overall separation of normal and anomalous regions. The shallower-layer constraint emphasizes difficult local responses to reduce spurious activations caused by normal textures. This assignment distinguishes supervision roles; each layer may still contain both local and global information.

Finally, we use \(\pi =\sigma (\mathbf{H})\) to compute soft true positives, false positives, and false negatives. A Tversky loss encourages region completeness in the fused anomaly map~\citep{salehi2017tversky}:

\begin{equation}\label{eq:cear-17}
\mathcal{L}_{\mathrm{map}}=\mathbb{E}_{\mathbf{x}}[1-\frac{\operatorname{TP}+\epsilon }{\operatorname{TP}+a_{T}\operatorname{FP}+b_{T}\operatorname{FN}+\epsilon }].
\end{equation}

The coefficients \(a_{T},b_{T}\) control the relative costs of false positives and false negatives, respectively, while \(\epsilon \) provides numerical stability. The joint objective is

\begin{equation}\label{eq:cear-18}
\begin{aligned}
\mathcal{L}&=\lambda _{\mathrm{cls}}\mathcal{L}_{\mathrm{cls}}+\lambda _{\mathrm{seg}}\mathcal{L}_{\mathrm{seg}}+\lambda _{\mathrm{qry}}\mathcal{L}_{\mathrm{qry}} \\
&+\lambda _{\mathrm{sem}}\mathcal{L}_{\mathrm{sem}}+\lambda _{\mathrm{tex}}\mathcal{L}_{\mathrm{tex}}+\lambda _{\mathrm{map}}\mathcal{L}_{\mathrm{map}}.
\end{aligned}
\end{equation}

The coefficients \(\lambda \) weight the loss terms; their values and the training configuration are given in Section~\ref{sec4-1}. Anomaly adaptation updates the queries, projections, memory encoding, and conditional refinement modules, together with the shared layer normalization at the ViT output. The ViT backbone and CNN encoder remain frozen throughout anomaly adaptation.

\begin{table*}[!t]
\centering
\caption{Comparison of zero-shot anomaly detection methods on 13 industrial and medical datasets. Image-level metrics are AUROC, F1-max and AP; pixel-level metrics are AUROC, F1-max, AP and PRO. Best and second-best reported values for each dataset and metric are shown in \textcolor[HTML]{FF0000}{red} and \textcolor[HTML]{0000FF}{blue}.}
\label{tab1}
\begingroup
\fontfamily{ptm}\fontsize{7}{8.5}\selectfont
\setlength{\tabcolsep}{0.6pt}
\renewcommand{\arraystretch}{1.65}
\setlength{\aboverulesep}{0pt}
\setlength{\belowrulesep}{0pt}
\renewcommand{\tabularxcolumn}[1]{m{#1}}
\begin{tabularx}{\textwidth}{@{} >{\centering\arraybackslash}m{9pt} >{\centering\arraybackslash}m{40pt} >{\centering\arraybackslash}m{35pt} *{6}{>{\centering\arraybackslash}X} *{1}{>{\columncolor[HTML]{E7E7E7}\centering\arraybackslash}X} @{}}
\toprule[1pt]
\rule{0pt}{13pt}\multirow[c]{2}{=}{\centering\rotatebox[origin=c]{90}{Domain}} & \multirow[c]{2}{=}{\centering Metric} & \multirow[c]{2}{=}{\centering Dataset} & WinCLIP & APRIL-GAN & CLIP-AD & AnomalyCLIP & AdaCLIP & \shortstack[c]{VisualAD\\(CLIP)} & \textbf{CEAR} \\
\cmidrule[0.5pt]{4-10}
 & & & {\fontsize{6}{7}\selectfont ViT-L/14 @336px} & {\fontsize{6}{7}\selectfont ViT-L/14 @336px} & {\fontsize{6}{7}\selectfont ViT-L/14 @336px} & {\fontsize{6}{7}\selectfont ViT-L/14 @336px} & {\fontsize{6}{7}\selectfont ViT-L/14 @336px} & {\fontsize{6}{7}\selectfont ViT-L/14 @336px} & {\fontsize{6}{7}\selectfont\mbox{MoonViT @672px}} \\
\midrule[0.5pt]
\multirow[c]{10}{=}{\centering\rotatebox[origin=c]{90}{Industrial}} & \multirow[c]{5}{=}{\centering\shortstack[c]{Image-level\\(AUROC,\\F1-max, AP)}} & VisA & \mbox{(\makebox[15pt][r]{75.6},\hspace{0.6pt}\makebox[15pt][r]{78.2},\hspace{0.6pt}\makebox[15pt][r]{78.8})} & \mbox{(\makebox[15pt][r]{77.4},\hspace{0.6pt}\makebox[15pt][r]{78.6},\hspace{0.6pt}\makebox[15pt][r]{80.9})} & \mbox{(\makebox[15pt][r]{66.2},\hspace{0.6pt}\makebox[15pt][r]{74.3},\hspace{0.6pt}\makebox[15pt][r]{71.4})} & \mbox{(\makebox[15pt][r]{81.0},\hspace{0.6pt}\makebox[15pt][r]{80.3},\hspace{0.6pt}\makebox[15pt][r]{84.4})} & \mbox{(\makebox[15pt][r]{79.7},\hspace{0.6pt}\makebox[15pt][r]{79.6},\hspace{0.6pt}\makebox[15pt][r]{83.2})} & \mbox{(\makebox[15pt][r]{\textcolor[HTML]{0000FF}{84.7}},\hspace{0.6pt}\makebox[15pt][r]{\textcolor[HTML]{0000FF}{82.5}},\hspace{0.6pt}\makebox[15pt][r]{\textcolor[HTML]{0000FF}{87.6}})} & \mbox{(\makebox[15pt][r]{\textcolor[HTML]{FF0000}{84.8}},\hspace{0.6pt}\makebox[15pt][r]{\textcolor[HTML]{FF0000}{82.7}},\hspace{0.6pt}\makebox[15pt][r]{\textcolor[HTML]{FF0000}{88.5}})} \\
 &  & MVTec-AD & \mbox{(\makebox[15pt][r]{90.4},\hspace{0.6pt}\makebox[15pt][r]{\textcolor[HTML]{0000FF}{92.7}},\hspace{0.6pt}\makebox[15pt][r]{95.6})} & \mbox{(\makebox[15pt][r]{86.1},\hspace{0.6pt}\makebox[15pt][r]{90.4},\hspace{0.6pt}\makebox[15pt][r]{93.6})} & \mbox{(\makebox[15pt][r]{74.1},\hspace{0.6pt}\makebox[15pt][r]{86.3},\hspace{0.6pt}\makebox[15pt][r]{88.1})} & \mbox{(\makebox[15pt][r]{91.6},\hspace{0.6pt}\makebox[15pt][r]{\textcolor[HTML]{0000FF}{92.7}},\hspace{0.6pt}\makebox[15pt][r]{96.2})} & \mbox{(\makebox[15pt][r]{92.0},\hspace{0.6pt}\makebox[15pt][r]{\textcolor[HTML]{0000FF}{92.7}},\hspace{0.6pt}\makebox[15pt][r]{96.4})} & \mbox{(\makebox[15pt][r]{\textcolor[HTML]{0000FF}{92.2}},\hspace{0.6pt}\makebox[15pt][r]{\textcolor[HTML]{FF0000}{93.2}},\hspace{0.6pt}\makebox[15pt][r]{\textcolor[HTML]{FF0000}{96.7}})} & \mbox{(\makebox[15pt][r]{\textcolor[HTML]{FF0000}{92.4}},\hspace{0.6pt}\makebox[15pt][r]{\textcolor[HTML]{0000FF}{92.7}},\hspace{0.6pt}\makebox[15pt][r]{\textcolor[HTML]{0000FF}{96.5}})} \\
 &  & BTAD & \mbox{(\makebox[15pt][r]{68.2},\hspace{0.6pt}\makebox[15pt][r]{67.8},\hspace{0.6pt}\makebox[15pt][r]{70.9})} & \mbox{(\makebox[15pt][r]{73.7},\hspace{0.6pt}\makebox[15pt][r]{68.7},\hspace{0.6pt}\makebox[15pt][r]{69.9})} & \mbox{(\makebox[15pt][r]{66.7},\hspace{0.6pt}\makebox[15pt][r]{65.9},\hspace{0.6pt}\makebox[15pt][r]{67.3})} & \mbox{(\makebox[15pt][r]{88.7},\hspace{0.6pt}\makebox[15pt][r]{86.0},\hspace{0.6pt}\makebox[15pt][r]{90.6})} & \mbox{(\makebox[15pt][r]{\textcolor[HTML]{0000FF}{90.0}},\hspace{0.6pt}\makebox[15pt][r]{87.2},\hspace{0.6pt}\makebox[15pt][r]{91.5})} & \mbox{(\makebox[15pt][r]{\textcolor[HTML]{FF0000}{94.9}},\hspace{0.6pt}\makebox[15pt][r]{\textcolor[HTML]{FF0000}{93.9}},\hspace{0.6pt}\makebox[15pt][r]{\textcolor[HTML]{FF0000}{97.0}})} & \mbox{(\makebox[15pt][r]{\textcolor[HTML]{FF0000}{94.9}},\hspace{0.6pt}\makebox[15pt][r]{\textcolor[HTML]{0000FF}{87.9}},\hspace{0.6pt}\makebox[15pt][r]{\textcolor[HTML]{0000FF}{92.6}})} \\
 &  & DTD & \mbox{(\makebox[15pt][r]{95.1},\hspace{0.6pt}\makebox[15pt][r]{94.1},\hspace{0.6pt}\makebox[15pt][r]{97.7})} & \mbox{(\makebox[15pt][r]{85.5},\hspace{0.6pt}\makebox[15pt][r]{89.1},\hspace{0.6pt}\makebox[15pt][r]{94.0})} & \mbox{(\makebox[15pt][r]{75.1},\hspace{0.6pt}\makebox[15pt][r]{86.3},\hspace{0.6pt}\makebox[15pt][r]{88.0})} & \mbox{(\makebox[15pt][r]{93.7},\hspace{0.6pt}\makebox[15pt][r]{94.3},\hspace{0.6pt}\makebox[15pt][r]{97.4})} & \mbox{(\makebox[15pt][r]{92.1},\hspace{0.6pt}\makebox[15pt][r]{92.4},\hspace{0.6pt}\makebox[15pt][r]{96.3})} & \mbox{(\makebox[15pt][r]{\textcolor[HTML]{0000FF}{97.5}},\hspace{0.6pt}\makebox[15pt][r]{\textcolor[HTML]{0000FF}{96.6}},\hspace{0.6pt}\makebox[15pt][r]{\textcolor[HTML]{0000FF}{99.1}})} & \mbox{(\makebox[15pt][r]{\textcolor[HTML]{FF0000}{97.6}},\hspace{0.6pt}\makebox[15pt][r]{\textcolor[HTML]{FF0000}{97.4}},\hspace{0.6pt}\makebox[15pt][r]{\textcolor[HTML]{FF0000}{99.3}})} \\
 &  & MPDD & \mbox{(\makebox[15pt][r]{61.4},\hspace{0.6pt}\makebox[15pt][r]{77.5},\hspace{0.6pt}\makebox[15pt][r]{69.9})} & \mbox{(\makebox[15pt][r]{73.0},\hspace{0.6pt}\makebox[15pt][r]{76.0},\hspace{0.6pt}\makebox[15pt][r]{79.8})} & \mbox{(\makebox[15pt][r]{74.5},\hspace{0.6pt}\makebox[15pt][r]{72.9},\hspace{0.6pt}\makebox[15pt][r]{77.9})} & \mbox{(\makebox[15pt][r]{\textcolor[HTML]{0000FF}{77.0}},\hspace{0.6pt}\makebox[15pt][r]{75.4},\hspace{0.6pt}\makebox[15pt][r]{\textcolor[HTML]{FF0000}{82.0}})} & \mbox{(\makebox[15pt][r]{76.0},\hspace{0.6pt}\makebox[15pt][r]{\textcolor[HTML]{0000FF}{82.5}},\hspace{0.6pt}\makebox[15pt][r]{77.1})} & \mbox{(\makebox[15pt][r]{71.5},\hspace{0.6pt}\makebox[15pt][r]{81.6},\hspace{0.6pt}\makebox[15pt][r]{79.5})} & \mbox{(\makebox[15pt][r]{\textcolor[HTML]{FF0000}{78.2}},\hspace{0.6pt}\makebox[15pt][r]{\textcolor[HTML]{FF0000}{85.2}},\hspace{0.6pt}\makebox[15pt][r]{\textcolor[HTML]{0000FF}{80.2}})} \\
\cmidrule[0.5pt]{2-10}
 & \multirow[c]{5}{=}{\centering\shortstack[c]{Pixel-level\\(AUROC,\\F1-max,\\AP, PRO)}} & VisA & \mbox{(\makebox[12.3pt][r]{73.2},\hspace{0.6pt}\makebox[12.3pt][r]{9.0},\hspace{0.6pt}\makebox[12.3pt][r]{5.4},\hspace{0.6pt}\makebox[12.3pt][r]{51.1})} & \mbox{(\makebox[12.3pt][r]{93.8},\hspace{0.6pt}\makebox[12.3pt][r]{32.6},\hspace{0.6pt}\makebox[12.3pt][r]{26.2},\hspace{0.6pt}\makebox[12.3pt][r]{86.5})} & \mbox{(\makebox[12.3pt][r]{93.0},\hspace{0.6pt}\makebox[12.3pt][r]{24.1},\hspace{0.6pt}\makebox[12.3pt][r]{17.9},\hspace{0.6pt}\makebox[12.3pt][r]{80.2})} & \mbox{(\makebox[12.3pt][r]{95.4},\hspace{0.6pt}\makebox[12.3pt][r]{27.6},\hspace{0.6pt}\makebox[12.3pt][r]{20.7},\hspace{0.6pt}\makebox[12.3pt][r]{86.4})} & \mbox{(\makebox[12.3pt][r]{95.1},\hspace{0.6pt}\makebox[12.3pt][r]{33.8},\hspace{0.6pt}\makebox[12.3pt][r]{\textcolor[HTML]{FF0000}{29.2}},\hspace{0.6pt}\makebox[12.3pt][r]{71.3})} & \mbox{(\makebox[12.3pt][r]{\textcolor[HTML]{0000FF}{95.8}},\hspace{0.6pt}\makebox[12.3pt][r]{\textcolor[HTML]{0000FF}{34.6}},\hspace{0.6pt}\makebox[12.3pt][r]{\textcolor[HTML]{0000FF}{28.4}},\hspace{0.6pt}\makebox[12.3pt][r]{\textcolor[HTML]{FF0000}{91.0}})} & \mbox{(\makebox[12.3pt][r]{\textcolor[HTML]{FF0000}{95.9}},\hspace{0.6pt}\makebox[12.3pt][r]{\textcolor[HTML]{FF0000}{34.9}},\hspace{0.6pt}\makebox[12.3pt][r]{\textcolor[HTML]{0000FF}{28.4}},\hspace{0.6pt}\makebox[12.3pt][r]{\textcolor[HTML]{0000FF}{88.5}})} \\
 &  & MVTec-AD & \mbox{(\makebox[12.3pt][r]{82.3},\hspace{0.6pt}\makebox[12.3pt][r]{24.8},\hspace{0.6pt}\makebox[12.3pt][r]{18.2},\hspace{0.6pt}\makebox[12.3pt][r]{62.0})} & \mbox{(\makebox[12.3pt][r]{87.5},\hspace{0.6pt}\makebox[12.3pt][r]{42.3},\hspace{0.6pt}\makebox[12.3pt][r]{39.1},\hspace{0.6pt}\makebox[12.3pt][r]{43.7})} & \mbox{(\makebox[12.3pt][r]{77.9},\hspace{0.6pt}\makebox[12.3pt][r]{26.3},\hspace{0.6pt}\makebox[12.3pt][r]{21.1},\hspace{0.6pt}\makebox[12.3pt][r]{55.7})} & \mbox{(\makebox[12.3pt][r]{\textcolor[HTML]{0000FF}{91.0}},\hspace{0.6pt}\makebox[12.3pt][r]{38.9},\hspace{0.6pt}\makebox[12.3pt][r]{34.4},\hspace{0.6pt}\makebox[12.3pt][r]{81.7})} & \mbox{(\makebox[12.3pt][r]{88.5},\hspace{0.6pt}\makebox[12.3pt][r]{\textcolor[HTML]{0000FF}{43.9}},\hspace{0.6pt}\makebox[12.3pt][r]{41.0},\hspace{0.6pt}\makebox[12.3pt][r]{47.6})} & \mbox{(\makebox[12.3pt][r]{90.8},\hspace{0.6pt}\makebox[12.3pt][r]{\textcolor[HTML]{0000FF}{43.9}},\hspace{0.6pt}\makebox[12.3pt][r]{\textcolor[HTML]{0000FF}{41.2}},\hspace{0.6pt}\makebox[12.3pt][r]{\textcolor[HTML]{0000FF}{87.5}})} & \mbox{(\makebox[12.3pt][r]{\textcolor[HTML]{FF0000}{91.2}},\hspace{0.6pt}\makebox[12.3pt][r]{\textcolor[HTML]{FF0000}{46.3}},\hspace{0.6pt}\makebox[12.3pt][r]{\textcolor[HTML]{FF0000}{45.0}},\hspace{0.6pt}\makebox[12.3pt][r]{\textcolor[HTML]{FF0000}{87.9}})} \\
 &  & BTAD & \mbox{(\makebox[12.3pt][r]{72.7},\hspace{0.6pt}\makebox[12.3pt][r]{18.5},\hspace{0.6pt}\makebox[12.3pt][r]{12.9},\hspace{0.6pt}\makebox[12.3pt][r]{27.3})} & \mbox{(\makebox[12.3pt][r]{91.3},\hspace{0.6pt}\makebox[12.3pt][r]{40.1},\hspace{0.6pt}\makebox[12.3pt][r]{37.7},\hspace{0.6pt}\makebox[12.3pt][r]{21.0})} & \mbox{(\makebox[12.3pt][r]{80.9},\hspace{0.6pt}\makebox[12.3pt][r]{24.1},\hspace{0.6pt}\makebox[12.3pt][r]{18.3},\hspace{0.6pt}\makebox[12.3pt][r]{41.4})} & \mbox{(\makebox[12.3pt][r]{\textcolor[HTML]{0000FF}{93.0}},\hspace{0.6pt}\makebox[12.3pt][r]{47.1},\hspace{0.6pt}\makebox[12.3pt][r]{41.5},\hspace{0.6pt}\makebox[12.3pt][r]{71.0})} & \mbox{(\makebox[12.3pt][r]{87.7},\hspace{0.6pt}\makebox[12.3pt][r]{42.3},\hspace{0.6pt}\makebox[12.3pt][r]{36.6},\hspace{0.6pt}\makebox[12.3pt][r]{17.1})} & \mbox{(\makebox[12.3pt][r]{91.1},\hspace{0.6pt}\makebox[12.3pt][r]{\textcolor[HTML]{FF0000}{49.8}},\hspace{0.6pt}\makebox[12.3pt][r]{\textcolor[HTML]{0000FF}{43.1}},\hspace{0.6pt}\makebox[12.3pt][r]{\textcolor[HTML]{0000FF}{80.4}})} & \mbox{(\makebox[12.3pt][r]{\textcolor[HTML]{FF0000}{96.7}},\hspace{0.6pt}\makebox[12.3pt][r]{\textcolor[HTML]{0000FF}{49.3}},\hspace{0.6pt}\makebox[12.3pt][r]{\textcolor[HTML]{FF0000}{44.0}},\hspace{0.6pt}\makebox[12.3pt][r]{\textcolor[HTML]{FF0000}{84.9}})} \\
 &  & DTD & \mbox{(\makebox[12.3pt][r]{79.5},\hspace{0.6pt}\makebox[12.3pt][r]{16.1},\hspace{0.6pt}\makebox[12.3pt][r]{9.8},\hspace{0.6pt}\makebox[12.3pt][r]{51.5})} & \mbox{(\makebox[12.3pt][r]{94.9},\hspace{0.6pt}\makebox[12.3pt][r]{60.4},\hspace{0.6pt}\makebox[12.3pt][r]{61.0},\hspace{0.6pt}\makebox[12.3pt][r]{33.8})} & \mbox{(\makebox[12.3pt][r]{86.6},\hspace{0.6pt}\makebox[12.3pt][r]{35.8},\hspace{0.6pt}\makebox[12.3pt][r]{31.0},\hspace{0.6pt}\makebox[12.3pt][r]{63.2})} & \mbox{(\makebox[12.3pt][r]{97.5},\hspace{0.6pt}\makebox[12.3pt][r]{55.8},\hspace{0.6pt}\makebox[12.3pt][r]{52.5},\hspace{0.6pt}\makebox[12.3pt][r]{87.9})} & \mbox{(\makebox[12.3pt][r]{95.1},\hspace{0.6pt}\makebox[12.3pt][r]{58.4},\hspace{0.6pt}\makebox[12.3pt][r]{56.1},\hspace{0.6pt}\makebox[12.3pt][r]{34.3})} & \mbox{(\makebox[12.3pt][r]{\textcolor[HTML]{0000FF}{98.1}},\hspace{0.6pt}\makebox[12.3pt][r]{\textcolor[HTML]{0000FF}{64.3}},\hspace{0.6pt}\makebox[12.3pt][r]{\textcolor[HTML]{0000FF}{65.5}},\hspace{0.6pt}\makebox[12.3pt][r]{\textcolor[HTML]{0000FF}{94.8}})} & \mbox{(\makebox[12.3pt][r]{\textcolor[HTML]{FF0000}{98.7}},\hspace{0.6pt}\makebox[12.3pt][r]{\textcolor[HTML]{FF0000}{66.6}},\hspace{0.6pt}\makebox[12.3pt][r]{\textcolor[HTML]{FF0000}{70.0}},\hspace{0.6pt}\makebox[12.3pt][r]{\textcolor[HTML]{FF0000}{95.5}})} \\
 &  & MPDD & \mbox{(\makebox[12.3pt][r]{71.2},\hspace{0.6pt}\makebox[12.3pt][r]{15.4},\hspace{0.6pt}\makebox[12.3pt][r]{22.6},\hspace{0.6pt}\makebox[12.3pt][r]{48.9})} & \mbox{(\makebox[12.3pt][r]{95.1},\hspace{0.6pt}\makebox[12.3pt][r]{30.6},\hspace{0.6pt}\makebox[12.3pt][r]{28.9},\hspace{0.6pt}\makebox[12.3pt][r]{83.1})} & \mbox{(\makebox[12.3pt][r]{95.1},\hspace{0.6pt}\makebox[12.3pt][r]{26.3},\hspace{0.6pt}\makebox[12.3pt][r]{28.4},\hspace{0.6pt}\makebox[12.3pt][r]{41.5})} & \mbox{(\makebox[12.3pt][r]{96.5},\hspace{0.6pt}\makebox[12.3pt][r]{32.1},\hspace{0.6pt}\makebox[12.3pt][r]{29.5},\hspace{0.6pt}\makebox[12.3pt][r]{88.7})} & \mbox{(\makebox[12.3pt][r]{96.1},\hspace{0.6pt}\makebox[12.3pt][r]{34.9},\hspace{0.6pt}\makebox[12.3pt][r]{32.9},\hspace{0.6pt}\makebox[12.3pt][r]{82.0})} & \mbox{(\makebox[12.3pt][r]{\textcolor[HTML]{0000FF}{97.1}},\hspace{0.6pt}\makebox[12.3pt][r]{\textcolor[HTML]{0000FF}{36.0}},\hspace{0.6pt}\makebox[12.3pt][r]{\textcolor[HTML]{0000FF}{33.3}},\hspace{0.6pt}\makebox[12.3pt][r]{\textcolor[HTML]{0000FF}{91.4}})} & \mbox{(\makebox[12.3pt][r]{\textcolor[HTML]{FF0000}{97.4}},\hspace{0.6pt}\makebox[12.3pt][r]{\textcolor[HTML]{FF0000}{40.8}},\hspace{0.6pt}\makebox[12.3pt][r]{\textcolor[HTML]{FF0000}{35.5}},\hspace{0.6pt}\makebox[12.3pt][r]{\textcolor[HTML]{FF0000}{92.7}})} \\
\midrule[0.5pt]
\multirow[c]{9}{=}{\centering\rotatebox[origin=c]{90}{Medical}} & \multirow[c]{3}{=}{\centering\shortstack[c]{Image-level\\(AUROC,\\F1-max, AP)}} & COVID-19 & \mbox{(\makebox[15pt][r]{64.4},\hspace{0.6pt}\makebox[15pt][r]{42.5},\hspace{0.6pt}\makebox[15pt][r]{42.9})} & \mbox{(\makebox[15pt][r]{15.5},\hspace{0.6pt}\makebox[15pt][r]{33.7},\hspace{0.6pt}\makebox[15pt][r]{40.6})} & \mbox{(\makebox[15pt][r]{62.5},\hspace{0.6pt}\makebox[15pt][r]{56.3},\hspace{0.6pt}\makebox[15pt][r]{43.2})} & \mbox{(\makebox[15pt][r]{\textcolor[HTML]{FF0000}{80.1}},\hspace{0.6pt}\makebox[15pt][r]{53.6},\hspace{0.6pt}\makebox[15pt][r]{\textcolor[HTML]{0000FF}{58.7}})} & \mbox{(\makebox[15pt][r]{64.5},\hspace{0.6pt}\makebox[15pt][r]{49.2},\hspace{0.6pt}\makebox[15pt][r]{50.3})} & \mbox{(\makebox[15pt][r]{43.8},\hspace{0.6pt}\makebox[15pt][r]{\textcolor[HTML]{0000FF}{68.3}},\hspace{0.6pt}\makebox[15pt][r]{45.9})} & \mbox{(\makebox[15pt][r]{\textcolor[HTML]{0000FF}{64.8}},\hspace{0.6pt}\makebox[15pt][r]{\textcolor[HTML]{FF0000}{68.4}},\hspace{0.6pt}\makebox[15pt][r]{\textcolor[HTML]{FF0000}{64.4}})} \\
 &  & Br35H & \mbox{(\makebox[15pt][r]{80.5},\hspace{0.6pt}\makebox[15pt][r]{74.1},\hspace{0.6pt}\makebox[15pt][r]{82.2})} & \mbox{(\makebox[15pt][r]{93.1},\hspace{0.6pt}\makebox[15pt][r]{85.4},\hspace{0.6pt}\makebox[15pt][r]{92.9})} & \mbox{(\makebox[15pt][r]{96.0},\hspace{0.6pt}\makebox[15pt][r]{90.8},\hspace{0.6pt}\makebox[15pt][r]{95.5})} & \mbox{(\makebox[15pt][r]{\textcolor[HTML]{FF0000}{97.3}},\hspace{0.6pt}\makebox[15pt][r]{\textcolor[HTML]{0000FF}{92.4}},\hspace{0.6pt}\makebox[15pt][r]{\textcolor[HTML]{FF0000}{96.1}})} & \mbox{(\makebox[15pt][r]{95.7},\hspace{0.6pt}\makebox[15pt][r]{89.1},\hspace{0.6pt}\makebox[15pt][r]{\textcolor[HTML]{0000FF}{95.7}})} & \mbox{(\makebox[15pt][r]{94.4},\hspace{0.6pt}\makebox[15pt][r]{87.9},\hspace{0.6pt}\makebox[15pt][r]{91.6})} & \mbox{(\makebox[15pt][r]{\textcolor[HTML]{0000FF}{96.2}},\hspace{0.6pt}\makebox[15pt][r]{\textcolor[HTML]{FF0000}{93.3}},\hspace{0.6pt}\makebox[15pt][r]{93.2})} \\
 &  & Brain AD & \mbox{(\makebox[15pt][r]{72.5},\hspace{0.6pt}\makebox[15pt][r]{90.7},\hspace{0.6pt}\makebox[15pt][r]{91.6})} & \mbox{(\makebox[15pt][r]{58.8},\hspace{0.6pt}\makebox[15pt][r]{90.6},\hspace{0.6pt}\makebox[15pt][r]{87.8})} & \mbox{(\makebox[15pt][r]{72.1},\hspace{0.6pt}\makebox[15pt][r]{91.3},\hspace{0.6pt}\makebox[15pt][r]{91.5})} & \mbox{(\makebox[15pt][r]{69.0},\hspace{0.6pt}\makebox[15pt][r]{90.6},\hspace{0.6pt}\makebox[15pt][r]{90.1})} & \mbox{(\makebox[15pt][r]{80.0},\hspace{0.6pt}\makebox[15pt][r]{90.2},\hspace{0.6pt}\makebox[15pt][r]{94.1})} & \mbox{(\makebox[15pt][r]{\textcolor[HTML]{0000FF}{80.8}},\hspace{0.6pt}\makebox[15pt][r]{\textcolor[HTML]{0000FF}{91.8}},\hspace{0.6pt}\makebox[15pt][r]{\textcolor[HTML]{0000FF}{94.7}})} & \mbox{(\makebox[15pt][r]{\textcolor[HTML]{FF0000}{83.6}},\hspace{0.6pt}\makebox[15pt][r]{\textcolor[HTML]{FF0000}{92.6}},\hspace{0.6pt}\makebox[15pt][r]{\textcolor[HTML]{FF0000}{95.7}})} \\
\cmidrule[0.5pt]{2-10}
 & \multirow[c]{6}{=}{\centering\shortstack[c]{Pixel-level\\(AUROC,\\F1-max,\\AP, PRO)}} & ClinicDB & \mbox{(\makebox[12.3pt][r]{70.3},\hspace{0.6pt}\makebox[12.3pt][r]{27.3},\hspace{0.6pt}\makebox[12.3pt][r]{19.4},\hspace{0.6pt}\makebox[12.3pt][r]{32.5})} & \mbox{(\makebox[12.3pt][r]{82.4},\hspace{0.6pt}\makebox[12.3pt][r]{38.2},\hspace{0.6pt}\makebox[12.3pt][r]{36.4},\hspace{0.6pt}\makebox[12.3pt][r]{53.5})} & \mbox{(\makebox[12.3pt][r]{76.5},\hspace{0.6pt}\makebox[12.3pt][r]{32.1},\hspace{0.6pt}\makebox[12.3pt][r]{24.6},\hspace{0.6pt}\makebox[12.3pt][r]{52.8})} & \mbox{(\makebox[12.3pt][r]{84.6},\hspace{0.6pt}\makebox[12.3pt][r]{45.5},\hspace{0.6pt}\makebox[12.3pt][r]{41.7},\hspace{0.6pt}\makebox[12.3pt][r]{69.8})} & \mbox{(\makebox[12.3pt][r]{84.3},\hspace{0.6pt}\makebox[12.3pt][r]{44.4},\hspace{0.6pt}\makebox[12.3pt][r]{\textcolor[HTML]{0000FF}{43.3}},\hspace{0.6pt}\makebox[12.3pt][r]{63.6})} & \mbox{(\makebox[12.3pt][r]{\textcolor[HTML]{0000FF}{85.2}},\hspace{0.6pt}\makebox[12.3pt][r]{\textcolor[HTML]{0000FF}{45.6}},\hspace{0.6pt}\makebox[12.3pt][r]{37.8},\hspace{0.6pt}\makebox[12.3pt][r]{\textcolor[HTML]{0000FF}{70.6}})} & \mbox{(\makebox[12.3pt][r]{\textcolor[HTML]{FF0000}{88.9}},\hspace{0.6pt}\makebox[12.3pt][r]{\textcolor[HTML]{FF0000}{52.2}},\hspace{0.6pt}\makebox[12.3pt][r]{\textcolor[HTML]{FF0000}{44.9}},\hspace{0.6pt}\makebox[12.3pt][r]{\textcolor[HTML]{FF0000}{76.1}})} \\
 &  & Endo & \mbox{(\makebox[12.3pt][r]{68.2},\hspace{0.6pt}\makebox[12.3pt][r]{32.9},\hspace{0.6pt}\makebox[12.3pt][r]{23.8},\hspace{0.6pt}\makebox[12.3pt][r]{28.3})} & \mbox{(\makebox[12.3pt][r]{82.7},\hspace{0.6pt}\makebox[12.3pt][r]{46.9},\hspace{0.6pt}\makebox[12.3pt][r]{47.2},\hspace{0.6pt}\makebox[12.3pt][r]{48.3})} & \mbox{(\makebox[12.3pt][r]{78.1},\hspace{0.6pt}\makebox[12.3pt][r]{41.4},\hspace{0.6pt}\makebox[12.3pt][r]{36.8},\hspace{0.6pt}\makebox[12.3pt][r]{46.4})} & \mbox{(\makebox[12.3pt][r]{\textcolor[HTML]{0000FF}{86.5}},\hspace{0.6pt}\makebox[12.3pt][r]{\textcolor[HTML]{0000FF}{53.7}},\hspace{0.6pt}\makebox[12.3pt][r]{\textcolor[HTML]{0000FF}{50.7}},\hspace{0.6pt}\makebox[12.3pt][r]{\textcolor[HTML]{0000FF}{67.1}})} & \mbox{(\makebox[12.3pt][r]{82.9},\hspace{0.6pt}\makebox[12.3pt][r]{47.9},\hspace{0.6pt}\makebox[12.3pt][r]{47.7},\hspace{0.6pt}\makebox[12.3pt][r]{55.8})} & \mbox{(\makebox[12.3pt][r]{84.9},\hspace{0.6pt}\makebox[12.3pt][r]{52.0},\hspace{0.6pt}\makebox[12.3pt][r]{46.2},\hspace{0.6pt}\makebox[12.3pt][r]{66.3})} & \mbox{(\makebox[12.3pt][r]{\textcolor[HTML]{FF0000}{91.1}},\hspace{0.6pt}\makebox[12.3pt][r]{\textcolor[HTML]{FF0000}{62.2}},\hspace{0.6pt}\makebox[12.3pt][r]{\textcolor[HTML]{FF0000}{60.8}},\hspace{0.6pt}\makebox[12.3pt][r]{\textcolor[HTML]{FF0000}{76.7}})} \\
 &  & Kvasir & \mbox{(\makebox[12.3pt][r]{69.7},\hspace{0.6pt}\makebox[12.3pt][r]{35.9},\hspace{0.6pt}\makebox[12.3pt][r]{27.8},\hspace{0.6pt}\makebox[12.3pt][r]{24.5})} & \mbox{(\makebox[12.3pt][r]{77.6},\hspace{0.6pt}\makebox[12.3pt][r]{43.2},\hspace{0.6pt}\makebox[12.3pt][r]{42.5},\hspace{0.6pt}\makebox[12.3pt][r]{40.6})} & \mbox{(\makebox[12.3pt][r]{73.6},\hspace{0.6pt}\makebox[12.3pt][r]{38.9},\hspace{0.6pt}\makebox[12.3pt][r]{31.1},\hspace{0.6pt}\makebox[12.3pt][r]{30.6})} & \mbox{(\makebox[12.3pt][r]{\textcolor[HTML]{0000FF}{82.0}},\hspace{0.6pt}\makebox[12.3pt][r]{\textcolor[HTML]{0000FF}{50.0}},\hspace{0.6pt}\makebox[12.3pt][r]{43.2},\hspace{0.6pt}\makebox[12.3pt][r]{44.7})} & \mbox{(\makebox[12.3pt][r]{80.3},\hspace{0.6pt}\makebox[12.3pt][r]{47.7},\hspace{0.6pt}\makebox[12.3pt][r]{31.0},\hspace{0.6pt}\makebox[12.3pt][r]{27.5})} & \mbox{(\makebox[12.3pt][r]{80.3},\hspace{0.6pt}\makebox[12.3pt][r]{47.8},\hspace{0.6pt}\makebox[12.3pt][r]{\textcolor[HTML]{0000FF}{47.6}},\hspace{0.6pt}\makebox[12.3pt][r]{\textcolor[HTML]{0000FF}{50.8}})} & \mbox{(\makebox[12.3pt][r]{\textcolor[HTML]{FF0000}{88.5}},\hspace{0.6pt}\makebox[12.3pt][r]{\textcolor[HTML]{FF0000}{60.1}},\hspace{0.6pt}\makebox[12.3pt][r]{\textcolor[HTML]{FF0000}{56.2}},\hspace{0.6pt}\makebox[12.3pt][r]{\textcolor[HTML]{FF0000}{52.2}})} \\
 &  & ISIC & \mbox{(\makebox[12.3pt][r]{83.3},\hspace{0.6pt}\makebox[12.3pt][r]{53.4},\hspace{0.6pt}\makebox[12.3pt][r]{62.4},\hspace{0.6pt}\makebox[12.3pt][r]{55.1})} & \mbox{(\makebox[12.3pt][r]{85.8},\hspace{0.6pt}\makebox[12.3pt][r]{59.9},\hspace{0.6pt}\makebox[12.3pt][r]{69.8},\hspace{0.6pt}\makebox[12.3pt][r]{13.7})} & \mbox{(\makebox[12.3pt][r]{81.6},\hspace{0.6pt}\makebox[12.3pt][r]{63.6},\hspace{0.6pt}\makebox[12.3pt][r]{65.5},\hspace{0.6pt}\makebox[12.3pt][r]{29.0})} & \mbox{(\makebox[12.3pt][r]{\textcolor[HTML]{0000FF}{88.4}},\hspace{0.6pt}\makebox[12.3pt][r]{62.3},\hspace{0.6pt}\makebox[12.3pt][r]{\textcolor[HTML]{0000FF}{74.4}},\hspace{0.6pt}\makebox[12.3pt][r]{\textcolor[HTML]{0000FF}{78.1}})} & \mbox{(\makebox[12.3pt][r]{85.4},\hspace{0.6pt}\makebox[12.3pt][r]{61.9},\hspace{0.6pt}\makebox[12.3pt][r]{70.6},\hspace{0.6pt}\makebox[12.3pt][r]{5.3})} & \mbox{(\makebox[12.3pt][r]{84.0},\hspace{0.6pt}\makebox[12.3pt][r]{\textcolor[HTML]{0000FF}{65.1}},\hspace{0.6pt}\makebox[12.3pt][r]{67.7},\hspace{0.6pt}\makebox[12.3pt][r]{73.1})} & \mbox{(\makebox[12.3pt][r]{\textcolor[HTML]{FF0000}{89.4}},\hspace{0.6pt}\makebox[12.3pt][r]{\textcolor[HTML]{FF0000}{72.1}},\hspace{0.6pt}\makebox[12.3pt][r]{\textcolor[HTML]{FF0000}{75.7}},\hspace{0.6pt}\makebox[12.3pt][r]{\textcolor[HTML]{FF0000}{80.0}})} \\
 &  & TN3K & \mbox{(\makebox[12.3pt][r]{67.2},\hspace{0.6pt}\makebox[12.3pt][r]{30.0},\hspace{0.6pt}\makebox[12.3pt][r]{32.5},\hspace{0.6pt}\makebox[12.3pt][r]{39.8})} & \mbox{(\makebox[12.3pt][r]{75.9},\hspace{0.6pt}\makebox[12.3pt][r]{36.4},\hspace{0.6pt}\makebox[12.3pt][r]{35.4},\hspace{0.6pt}\makebox[12.3pt][r]{37.8})} & \mbox{(\makebox[12.3pt][r]{66.5},\hspace{0.6pt}\makebox[12.3pt][r]{32.8},\hspace{0.6pt}\makebox[12.3pt][r]{31.7},\hspace{0.6pt}\makebox[12.3pt][r]{35.9})} & \mbox{(\makebox[12.3pt][r]{76.8},\hspace{0.6pt}\makebox[12.3pt][r]{40.5},\hspace{0.6pt}\makebox[12.3pt][r]{\textcolor[HTML]{0000FF}{36.7}},\hspace{0.6pt}\makebox[12.3pt][r]{\textcolor[HTML]{0000FF}{50.3}})} & \mbox{(\makebox[12.3pt][r]{\textcolor[HTML]{0000FF}{77.2}},\hspace{0.6pt}\makebox[12.3pt][r]{\textcolor[HTML]{0000FF}{41.3}},\hspace{0.6pt}\makebox[12.3pt][r]{36.5},\hspace{0.6pt}\makebox[12.3pt][r]{17.0})} & \mbox{(\makebox[12.3pt][r]{65.5},\hspace{0.6pt}\makebox[12.3pt][r]{32.7},\hspace{0.6pt}\makebox[12.3pt][r]{30.4},\hspace{0.6pt}\makebox[12.3pt][r]{41.1})} & \mbox{(\makebox[12.3pt][r]{\textcolor[HTML]{FF0000}{79.1}},\hspace{0.6pt}\makebox[12.3pt][r]{\textcolor[HTML]{FF0000}{42.0}},\hspace{0.6pt}\makebox[12.3pt][r]{\textcolor[HTML]{FF0000}{36.9}},\hspace{0.6pt}\makebox[12.3pt][r]{\textcolor[HTML]{FF0000}{51.4}})} \\
 &  & Brain AD & \mbox{(\makebox[12.3pt][r]{87.6},\hspace{0.6pt}\makebox[12.3pt][r]{21.7},\hspace{0.6pt}\makebox[12.3pt][r]{13.3},\hspace{0.6pt}\makebox[12.3pt][r]{59.7})} & \mbox{(\makebox[12.3pt][r]{83.6},\hspace{0.6pt}\makebox[12.3pt][r]{38.7},\hspace{0.6pt}\makebox[12.3pt][r]{35.0},\hspace{0.6pt}\makebox[12.3pt][r]{45.4})} & \mbox{(\makebox[12.3pt][r]{94.1},\hspace{0.6pt}\makebox[12.3pt][r]{42.9},\hspace{0.6pt}\makebox[12.3pt][r]{40.5},\hspace{0.6pt}\makebox[12.3pt][r]{74.2})} & \mbox{(\makebox[12.3pt][r]{95.1},\hspace{0.6pt}\makebox[12.3pt][r]{43.1},\hspace{0.6pt}\makebox[12.3pt][r]{42.3},\hspace{0.6pt}\makebox[12.3pt][r]{71.5})} & \mbox{(\makebox[12.3pt][r]{\textcolor[HTML]{0000FF}{95.2}},\hspace{0.6pt}\makebox[12.3pt][r]{40.6},\hspace{0.6pt}\makebox[12.3pt][r]{37.0},\hspace{0.6pt}\makebox[12.3pt][r]{36.4})} & \mbox{(\makebox[12.3pt][r]{\textcolor[HTML]{0000FF}{95.2}},\hspace{0.6pt}\makebox[12.3pt][r]{\textcolor[HTML]{0000FF}{46.7}},\hspace{0.6pt}\makebox[12.3pt][r]{\textcolor[HTML]{0000FF}{43.7}},\hspace{0.6pt}\makebox[12.3pt][r]{\textcolor[HTML]{0000FF}{79.5}})} & \mbox{(\makebox[12.3pt][r]{\textcolor[HTML]{FF0000}{96.8}},\hspace{0.6pt}\makebox[12.3pt][r]{\textcolor[HTML]{FF0000}{53.5}},\hspace{0.6pt}\makebox[12.3pt][r]{\textcolor[HTML]{FF0000}{53.0}},\hspace{0.6pt}\makebox[12.3pt][r]{\textcolor[HTML]{FF0000}{81.9}})} \\
\bottomrule[1pt]
\end{tabularx}
\endgroup
\end{table*}

\section{Experiments}\label{sec4}

We evaluate how CEAR generalizes across industrial categories and medical imaging domains, then examine the contribution of cross-encoder anchor refinement under controlled conditions. The ablations isolate input resolution, industrial pretraining, multiscale memory, conditional interactions, and supervision at different layers. Qualitative analyses complement these comparisons and help characterize the model's range of applicability.

\subsection{Experimental Setup}\label{sec4-1}

\subsubsection{Datasets and Training Protocol}\label{sec4-1-1}

We evaluate anomaly detection and localization on 13 industrial and medical datasets. The industrial benchmarks are MVTec AD~\citep{bergmann2019mvtec}, VisA~\citep{zou2022spot}, BTAD~\citep{mishra2021vt}, DTD-Synthetic~\citep{aota2023zero}, and MPDD~\citep{jezek2021deep}. All five support image-level detection and pixel-level localization. MVTec AD and VisA alternate as auxiliary and target domains in the two training directions. DTD-Synthetic denotes the texture benchmark with synthetic anomalies, rather than the original texture classification task.

Medical image-level detection is evaluated on COVID-19 Radiography, Br35H, and Brain AD. COVID-19 Radiography is abbreviated as COVID-19 in the tables~\citep{chowdhury2020can,rahman2021exploring}. Br35H follows \citet{hamada2020br35h}, and Brain AD is a two-dimensional evaluation set derived from BraTS~\citep{baid2021rsna}. Medical localization uses CVC-ClinicDB~\citep{bernal2015wm}, Endo (the EndoTect segmentation benchmark~\citep{hicks2021endotect}), and Kvasir (Kvasir-SEG~\citep{jha2019kvasir}). It also includes ISIC (ISIC 2016~\citep{gutman2016skin}), TN3K~\citep{gong2021multi}, and Brain AD. These names follow the established anomaly detection benchmarks. The original datasets identify the data sources, whereas their anomaly detection subsets define the evaluation protocols. We therefore do not use sample counts from the original three-dimensional segmentation tasks or other dataset versions to describe the evaluated subsets.

For evaluation on VisA, we train on MVTec AD as the auxiliary dataset; all other target datasets are evaluated using the model trained on VisA. Auxiliary training uses normal/anomalous image labels and pixel masks, with one shared model per training direction. Target categories are excluded from anomaly adaptation, and inference requires neither target-domain normal reference images nor parameter updates. The CNN is separately pretrained for product-category recognition on Real-IAD, then frozen throughout anomaly adaptation and testing~\citep{wang2024real}.

\subsubsection{Evaluation Metrics}\label{sec4-1-2}

We measure image-level performance using AUROC, maximum F1 score (F1-max), and average precision (AP). Pixel-level evaluation additionally uses the area under the per-region overlap curve (PRO). For datasets with multiple categories, metrics are computed within each category and then macro-averaged. The controlled evaluation protocol uses three random seeds and summarizes the category-macro-averaged results by their mean and standard deviation. Comparisons are made separately for each metric.

\subsubsection{Implementation Details}\label{sec4-1-3}

We use MoonViT-SO-400M as the visual backbone, drawing on its multilayer patch representations and resolution flexibility for anomaly modeling at the encoder output~\citep{team2025kimi}. Inputs are resized to 672 \(\times\) 672 pixels. A patch size of 14 yields a 48 \(\times\) 48 grid with 1,152 channels. We extract features from layers \{7, 14, 21, 27\} of the 27-layer encoder. The shallowest branch receives supervision on difficult local regions; the other three form the deeper branch set for segmentation supervision. During anomaly adaptation, both encoder bodies remain frozen, while the shared final layer normalization of MoonViT and the anomaly modeling modules are updated. MoonViT inputs are normalized with a mean and standard deviation of 0.5; the CNN retains the normalization used during its pretraining.

The local encoder is ConvNeXt V2-Base~\citep{woo2023convnext}. We extract features at strides 4, 8, and 16, with 128, 256, and 512 channels, respectively. Each scale is projected to 256 channels using a 1 \(\times\) 1 convolution, group normalization with 32 groups, and GELU. A 3 \(\times\) 3 convolution with stride 2 downsamples the finest scale, while bilinear interpolation upsamples the coarsest scale to a common 84 \(\times\) 84 grid. The fusion mapping uses two 1 \(\times\) 1 convolutions, group normalization, and GELU. Its final layer is zero-initialized, with residual scaling \(\beta _{f}=0.1\). A bias-free memory projection produces 7,056 tokens of dimension 1,152.

Each selected layer uses four spatial anchors, and both memory encoding and conditional refinement use eight attention heads. The refinement feed-forward network has a hidden width of 4,608. Dropout is disabled in memory self-attention and set to 0.1 in conditional refinement and patch projection. The prototype residual coefficient \(\alpha _{\ell }\) is initialized to 0.01. Refined anchors pass directly to query-gated aggregation; all three refinement sublayers retain their internal residual connections. No additional pre-normalization is applied to the encoded memory.

The query margin \(\tau \) is set to 0.5. For the deep region loss, the foreground margin, background margin, background weight, and hard-example fraction are 0.5, 0.5, 0.25, and 1, respectively. The corresponding shallow-layer values are 0.5, 0.7, 1, and 0.25. The focal loss uses a focusing parameter of 2, and the Dice smoothing constant is 1. The Tversky loss uses \(a_{T}=0.3\), \(b_{T}=0.7\), and \(\epsilon =10^{-6}\). Loss weights for classification, segmentation, query separation, deep regions, shallow regions, and the fused map are 1, 1, 1, 0.5, 0.5, and 0.25, respectively.

Anomaly adaptation uses AdamW~\citep{loshchilov2017decoupled} for three epochs with a batch size of 4. The base learning rate is \(10^{-4}\), with weight decay of 0.01. The shared final layer normalization uses a learning rate of \(10^{-5}\) and no weight decay. An epoch-wise cosine schedule reduces the learning rate to a minimum of 0.1 times its initial value. Gradient norms are clipped at 0.5. The image score is the mean of the highest 1\% of anomaly-map responses. Pixel-level evaluation applies Gaussian smoothing with a standard deviation of 4; image scores are computed before smoothing.

Industrial pretraining fine-tunes a generically pretrained ConvNeXt V2-Base to recognize the 30 product categories in Real-IAD. Normal/anomalous labels and pixel masks do not enter this classification objective. Training uses 384 \(\times\) 384 inputs for 10 epochs, with 16 images per step and gradients accumulated over four steps. The AdamW learning rate is scaled by the effective batch size to 1.5625 \(\times\) \(10^{-4}\). Weight decay is 0.05, and the layer-wise learning-rate decay factor is 0.6. Augmentations comprise RandAugment, random erasing, Mixup, and CutMix~\citep{cubuk2020randaugment,zhong2020random,zhang2017mixup,yun2019cutmix}. The random-erasing probability is 0.25; the Beta-distribution parameters for Mixup and CutMix are 0.8 and 1.0, respectively. Label smoothing and stochastic depth are both set to 0.1, with an exponential moving average decay of 0.9999.

\subsection{Comparison with State-of-the-Art Methods}\label{sec4-2}

\subsubsection{Industrial Anomaly Detection and Localization}\label{sec4-2-1}

Table~\ref{tab1} compares CEAR with WinCLIP~\citep{jeong2023winclip}, APRIL-GAN~\citep{chen2023april}, and CLIP-AD~\citep{chen2024clip}. It also includes AnomalyCLIP~\citep{zhou2024anomalyclip}, AdaCLIP~\citep{cao2024adaclip}, and VisualAD~\citep{hou2026visualad}. This comparison evaluates each complete system. Subsequent ablations on VisA use matched backbones to isolate input resolution, industrial pretraining, and the interaction mechanism. Across 68 dataset--metric comparisons on 13 datasets, CEAR achieves 57 best results: 56 outright and one tied. Figure~\ref{fig3} relates these aggregate results to localization behavior through anomaly maps from industrial and medical target domains.

CEAR shows its strongest advantage in pixel-level localization, exceeding every other method in 41 of 44 comparisons and achieving the highest AUROC on all 11 localization datasets. Relative to the strongest competing result for each dataset and metric, mean gains in pixel AUROC, F1-max, AP, and PRO are 2.20, 4.45, 3.79, and 2.39 percentage points. These averages weight all 11 datasets equally and include comparisons in which CEAR does not rank first. CEAR ranks first in 17 of the 20 industrial localization comparisons. Pixel AP reaches 45.0 on MVTec AD, exceeding the strongest baseline by 3.8 percentage points. On MPDD, pixel F1-max reaches 40.8, a gain of 4.8 points; DTD-Synthetic pixel AP reaches 70.0, a gain of 4.5 points. The improvements span diverse industrial appearances and defect types, with complementary gains in pixel ranking, regional detection, and precision--recall performance.

\subsubsection{Transfer to Medical Imaging}\label{sec4-2-2}

CEAR achieves the highest scores in all 24 pixel-level comparisons across the six medical localization datasets. On Endo, pixel AP increases from the strongest baseline value of 50.7 to 60.8, while F1-max increases from 53.7 to 62.2. Pixel F1-max rises from 50.0 to 60.1 on Kvasir, and pixel AUROC rises from 77.2 to 79.1 on TN3K. Under the protocol in Section~\ref{sec4-1-1}, all medical datasets use the same model trained on VisA, without target-domain parameter updates. The localization gains therefore extend beyond industrial categories to brain, thyroid, skin, and endoscopic images. These results support cross-domain localization with industrial auxiliary supervision; the advantage is most pronounced at the pixel level.

\begin{figure*}[!t]
\centering
\includegraphics[width=\textwidth,height=0.77\textheight,keepaspectratio]{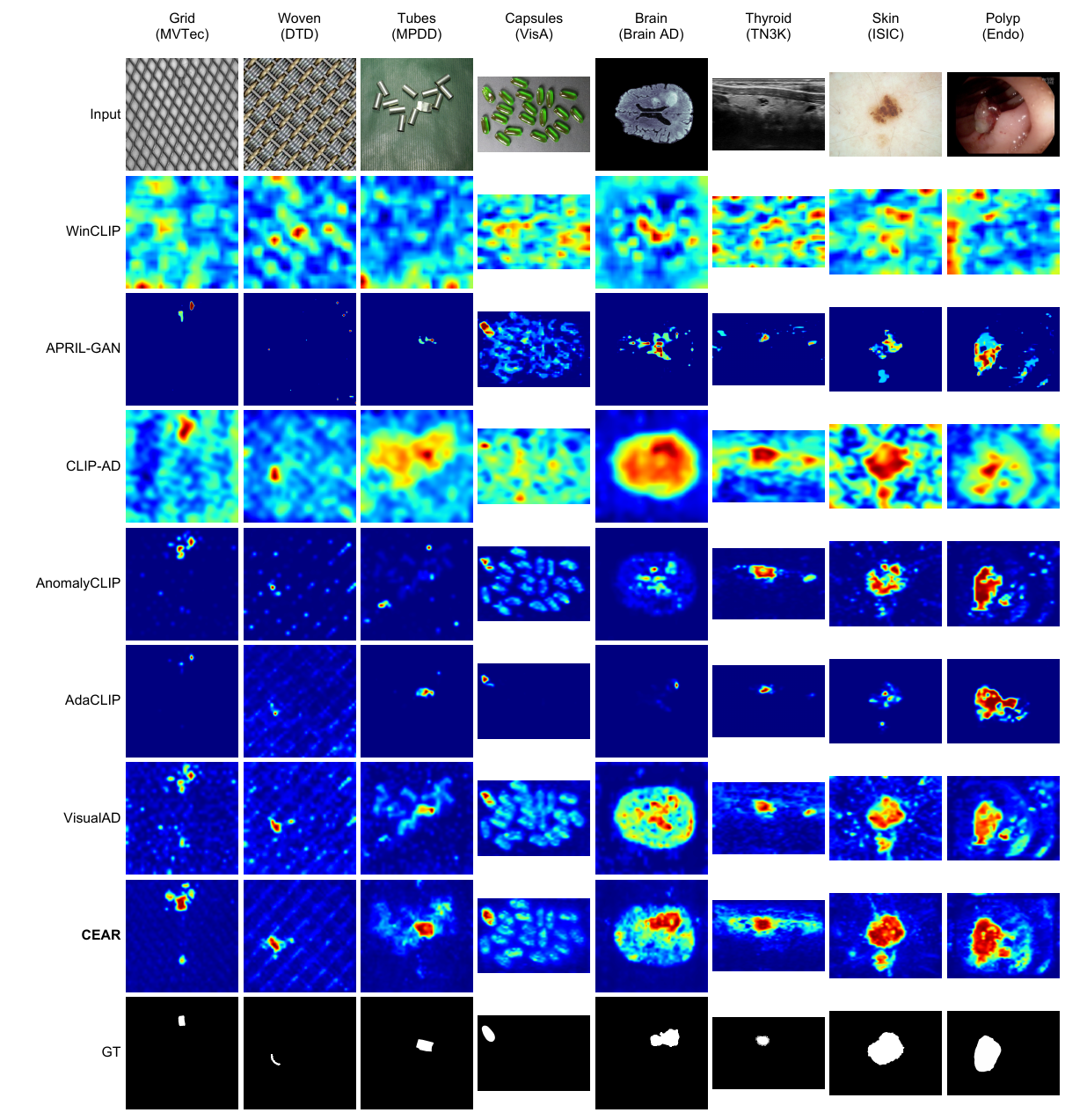}
\caption{Qualitative comparisons assess anomaly localization across diverse industrial and medical target domains. Columns show grid, woven-texture, tube, capsule, brain, thyroid, skin, and polyp examples, with input images, seven methods' anomaly maps, and ground-truth masks arranged vertically. Compared with several competing methods, CEAR shows closer spatial correspondence to annotated defects and lesions, with fewer fragmented responses or activations outside the annotated regions.}\label{fig3}
\end{figure*}

Figure~\ref{fig3} compares localization on eight industrial and medical sample types represented in Table~\ref{tab1}. For the illustrated grid, woven-texture, and metal-tube samples, CEAR retains localized responses at the annotated defects. Its high-response regions also show close spatial correspondence with the annotated lesions in brain, thyroid, skin, and polyp images. Other methods exhibit scattered activations, incomplete responses, or responses extending beyond annotated regions in these examples. The comparisons illustrate how CEAR organizes anomaly evidence across different appearances and imaging conditions. Table~\ref{tab1} quantifies aggregate performance, while Figure~\ref{fig3} shows the corresponding response locations and regional coverage. Heatmaps are normalized individually, so qualitative comparisons concern spatial distribution and correspondence with the ground truth.

\subsection{Ablation Studies}\label{sec4-3}

We conduct ablations on VisA~\citep{zou2022spot}, following the MVTec AD-to-VisA protocol in Section~\ref{sec4-1-1}. VisA target categories are excluded from anomaly adaptation. All configurations share the training split, optimization settings, and evaluation procedure, except for the factor under study. The default input resolution is 672 \(\times\) 672. Tables~\ref{tab2}--\ref{tab7} report all seven image-level and pixel-level metrics used in Table~\ref{tab1}.

\subsubsection{Backbone Resolution and the Contribution of Local Evidence}\label{sec4-3-1}

Table~\ref{tab2} compares controlled variants using the same MoonViT backbone at input resolutions of 448 \(\times\) 448, 560 \(\times\) 560, and 672 \(\times\) 672. MoonViT-only retains image-conditioned queries, spatial readout, patch projection, and the default supervision, while removing the CNN conditioning branch. The CNN-only controls include a fixed-feature response probe and a learned discriminative head trained with the same auxiliary anomaly supervision. Score-level fusion, feature concatenation, and anchor refinement before prototype construction use the same training split and optimization settings. These controls distinguish the effects of spatial resolution, additional local representations, and the stage at which the two encoding paths interact.

\begin{table}[!t]
\centering
\caption{Ablation of input resolution and local-evidence integration on VisA, comparing single-encoder models, score fusion, feature concatenation, and CEAR. \textbf{Bold} and \underline{underlined} values indicate the best and second-best results for each metric, respectively.}
\label{tab2}
\small
\setlength{\tabcolsep}{3pt}
\renewcommand{\arraystretch}{1.15}
\begin{adjustbox}{max width=\columnwidth}
\begin{tabular}{@{} l @{\hspace{8pt}} c *{7}{r} @{}}
\toprule
\multirow{2}{*}{Model} & \multirow{2}{*}{Input} & \multicolumn{4}{c}{Pixel-level} & \multicolumn{3}{c}{Image-level} \\
\cmidrule(lr){3-6}\cmidrule(l){7-9}
 & & \multicolumn{1}{c}{AUROC} & \multicolumn{1}{c}{F1-max} & \multicolumn{1}{c}{AP} & \multicolumn{1}{c}{PRO} & \multicolumn{1}{c}{AUROC} & \multicolumn{1}{c}{F1-max} & \multicolumn{1}{c}{AP} \\
\midrule
MoonViT-only & 448 & 95.2 & 31.5 & 25.9 & 85.0 & 79.5 & 78.5 & 84.1 \\
MoonViT-only & 560 & 95.3 & 31.9 & 26.7 & 85.4 & 81.5 & 80.0 & 85.5 \\
MoonViT-only & 672 & \underline{95.5} & \underline{33.3} & \underline{27.1} & \underline{87.5} & 82.3 & 80.1 & 85.9 \\
CNN-only (energy) & 672 & 80.8 & 13.2 & 8.4 & 60.4 & 64.9 & 73.6 & 68.4 \\
CNN-only (head) & 672 & 90.1 & 24.6 & 18.4 & 85.0 & 83.2 & 81.2 & 86.1 \\
Score fusion & 672 & 95.2 & 32.9 & 25.3 & 87.0 & 80.2 & 79.7 & 84.8 \\
Feature concat. & 672 & 95.2 & 32.8 & 25.7 & \underline{87.5} & 82.9 & 80.6 & 86.4 \\
CEAR & 448 & 95.4 & 31.9 & 26.0 & 86.9 & 82.2 & 80.9 & 85.9 \\
CEAR & 560 & \underline{95.5} & 32.2 & \underline{27.1} & 87.1 & \underline{83.3} & \underline{81.3} & \underline{87.1} \\
\textbf{CEAR (full)} & 672 & \textbf{95.9} & \textbf{34.9} & \textbf{28.4} & \textbf{88.5} & \textbf{84.8} & \textbf{82.7} & \textbf{88.5} \\
\bottomrule
\end{tabular}
\end{adjustbox}
\end{table}

CEAR improves all four pixel-level metrics over MoonViT-only at each of the three input resolutions. At 672 \(\times\) 672, gains in pixel AUROC, F1-max, AP, and PRO are 0.4, 1.6, 1.3, and 1.0 percentage points, respectively. Image AUROC, F1-max, and AP improve by 2.5, 2.6, and 2.6 points. The full model also outperforms score-level late fusion and direct feature concatenation. Thus, performance depends on how local evidence enters prototype construction, beyond the addition of a local encoding path. The learned CNN-only head is competitive on some image-level metrics, but its localization remains weaker than that of MoonViT-only. Using CNN features as conditioning memory for anchor refinement yields the best result on every metric in Table~\ref{tab2}. This comparison supports the joint use of local evidence and contextual representations.

\begin{figure}[!t]
\centering
\includegraphics[width=\columnwidth,height=0.79\textheight,keepaspectratio]{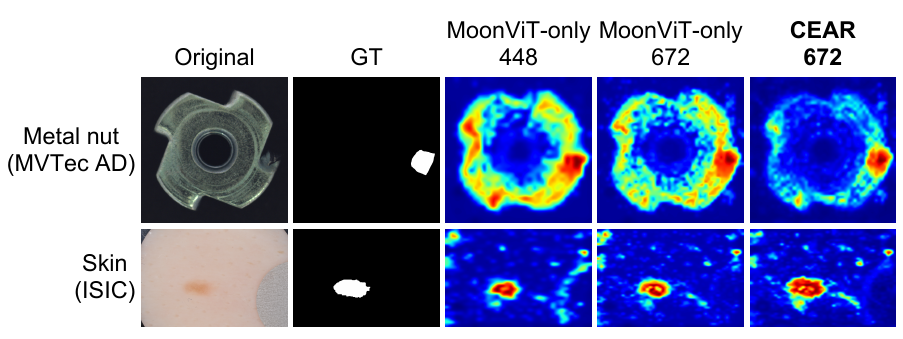}
\caption{Qualitative ablations distinguish the effects of increased input resolution and cross-encoder conditioning on anomaly localization. At matched resolution, CEAR reduces responses on normal ring structures and covers more of the annotated skin lesion.}\label{fig4}
\end{figure}

Figure~\ref{fig4} complements Table~\ref{tab2} by comparing MoonViT-only at resolutions of 448 and 672 with CEAR at 672 on the same images. For the MVTec AD metal-nut sample, increasing the MoonViT-only resolution leaves broad responses on the normal ring structure. At the same 672 resolution, CEAR concentrates high responses on the annotated defect to the right and suppresses some activations on normal structures. For the ISIC sample, CEAR covers more of the annotated lesion, whereas both single-backbone configurations concentrate on its center. These examples show distinct changes in response distribution from denser spatial sampling and conditional interaction. The controlled comparisons in Table~\ref{tab2} provide the quantitative evidence for the contribution of the interaction mechanism.

\subsubsection{Industrial Pretraining and Multiscale Memory}\label{sec4-3-2}

Table~\ref{tab3} varies the initialization of the frozen CNN: random initialization, generic image pretraining, or Real-IAD product-recognition pretraining. Memory construction and subsequent interactions remain unchanged. Table~\ref{tab4} compares single-scale features, pairs of scales, and the full feature pyramid. All variants are aligned to the same memory grid to control the token count.

\begin{table}[!t]
\centering
\caption{Ablation of CNN pretraining. Real-IAD pretraining uses product-category supervision; all CNN variants remain frozen during anomaly adaptation. Boldface and underlining follow the convention used in Table~\ref{tab2}.}
\label{tab3}
\small
\setlength{\tabcolsep}{3pt}
\renewcommand{\arraystretch}{1.15}
\begin{adjustbox}{max width=\columnwidth}
\begin{tabular}{@{} l @{\hspace{10pt}} c @{\hspace{14pt}} c @{}}
\toprule
Configuration & \shortstack{Pixel-level\\(AUROC, F1-max, AP, PRO)} & \shortstack{Image-level\\(AUROC, F1-max, AP)} \\
\midrule
Random init. & (\makebox[2.3em][r]{95.4}, \makebox[2.3em][r]{33.5}, \makebox[2.3em][r]{\underline{27.5}}, \makebox[2.3em][r]{\underline{88.1}}) & (\makebox[2.3em][r]{\underline{84.2}}, \makebox[2.3em][r]{82.0}, \makebox[2.3em][r]{\underline{87.7}}) \\
Generic pretraining & (\makebox[2.3em][r]{\underline{95.6}}, \makebox[2.3em][r]{\underline{34.0}}, \makebox[2.3em][r]{26.9}, \makebox[2.3em][r]{87.8}) & (\makebox[2.3em][r]{84.0}, \makebox[2.3em][r]{\underline{82.2}}, \makebox[2.3em][r]{87.5}) \\
\textbf{Real-IAD pretraining} & (\makebox[2.3em][r]{\textbf{95.9}}, \makebox[2.3em][r]{\textbf{34.9}}, \makebox[2.3em][r]{\textbf{28.4}}, \makebox[2.3em][r]{\textbf{88.5}}) & (\makebox[2.3em][r]{\textbf{84.8}}, \makebox[2.3em][r]{\textbf{82.7}}, \makebox[2.3em][r]{\textbf{88.5}}) \\
\bottomrule
\end{tabular}
\end{adjustbox}
\end{table}

\begin{table}[!t]
\centering
\caption{Ablation of multiscale local memory. C4, C8, and C16 denote CNN features at strides 4, 8, and 16, respectively. Boldface and underlining follow the convention used in Table~\ref{tab2}.}
\label{tab4}
\small
\setlength{\tabcolsep}{3pt}
\renewcommand{\arraystretch}{1.15}
\begin{adjustbox}{max width=\columnwidth}
\begin{tabular}{@{} l @{\hspace{10pt}} c @{\hspace{14pt}} c @{}}
\toprule
Configuration & \shortstack{Pixel-level\\(AUROC, F1-max, AP, PRO)} & \shortstack{Image-level\\(AUROC, F1-max, AP)} \\
\midrule
C4 & (\makebox[2.3em][r]{95.2}, \makebox[2.3em][r]{33.7}, \makebox[2.3em][r]{26.7}, \makebox[2.3em][r]{87.9}) & (\makebox[2.3em][r]{83.8}, \makebox[2.3em][r]{\underline{82.4}}, \makebox[2.3em][r]{87.3}) \\
C8 & (\makebox[2.3em][r]{\underline{95.5}}, \makebox[2.3em][r]{\underline{34.5}}, \makebox[2.3em][r]{\underline{27.5}}, \makebox[2.3em][r]{\underline{88.2}}) & (\makebox[2.3em][r]{\underline{84.4}}, \makebox[2.3em][r]{\underline{82.4}}, \makebox[2.3em][r]{\underline{87.8}}) \\
C16 & (\makebox[2.3em][r]{95.4}, \makebox[2.3em][r]{34.3}, \makebox[2.3em][r]{27.0}, \makebox[2.3em][r]{87.3}) & (\makebox[2.3em][r]{84.0}, \makebox[2.3em][r]{82.3}, \makebox[2.3em][r]{87.6}) \\
C4 + C8 & (\makebox[2.3em][r]{\underline{95.5}}, \makebox[2.3em][r]{34.1}, \makebox[2.3em][r]{27.0}, \makebox[2.3em][r]{87.4}) & (\makebox[2.3em][r]{84.1}, \makebox[2.3em][r]{82.1}, \makebox[2.3em][r]{87.4}) \\
C8 + C16 & (\makebox[2.3em][r]{\underline{95.5}}, \makebox[2.3em][r]{34.4}, \makebox[2.3em][r]{27.2}, \makebox[2.3em][r]{87.5}) & (\makebox[2.3em][r]{84.1}, \makebox[2.3em][r]{81.9}, \makebox[2.3em][r]{87.7}) \\
\textbf{C4 + C8 + C16} & (\makebox[2.3em][r]{\textbf{95.9}}, \makebox[2.3em][r]{\textbf{34.9}}, \makebox[2.3em][r]{\textbf{28.4}}, \makebox[2.3em][r]{\textbf{88.5}}) & (\makebox[2.3em][r]{\textbf{84.8}}, \makebox[2.3em][r]{\textbf{82.7}}, \makebox[2.3em][r]{\textbf{88.5}}) \\
\bottomrule
\end{tabular}
\end{adjustbox}
\end{table}

Industrial pretraining improves every localization metric over both random initialization and generic pretraining. The source of learned local representations therefore affects the usefulness of the conditioning memory. The finest scale preserves detail relevant to small defects but can also respond to normal texture; the coarsest scale provides broader component context. Combining these receptive fields in the full pyramid yields the strongest localization performance in Table~\ref{tab4}, consistent with complementary contributions to defect coverage and background suppression.

\begin{figure*}[!t]
\centering
\includegraphics[width=\textwidth,height=0.84\textheight,keepaspectratio]{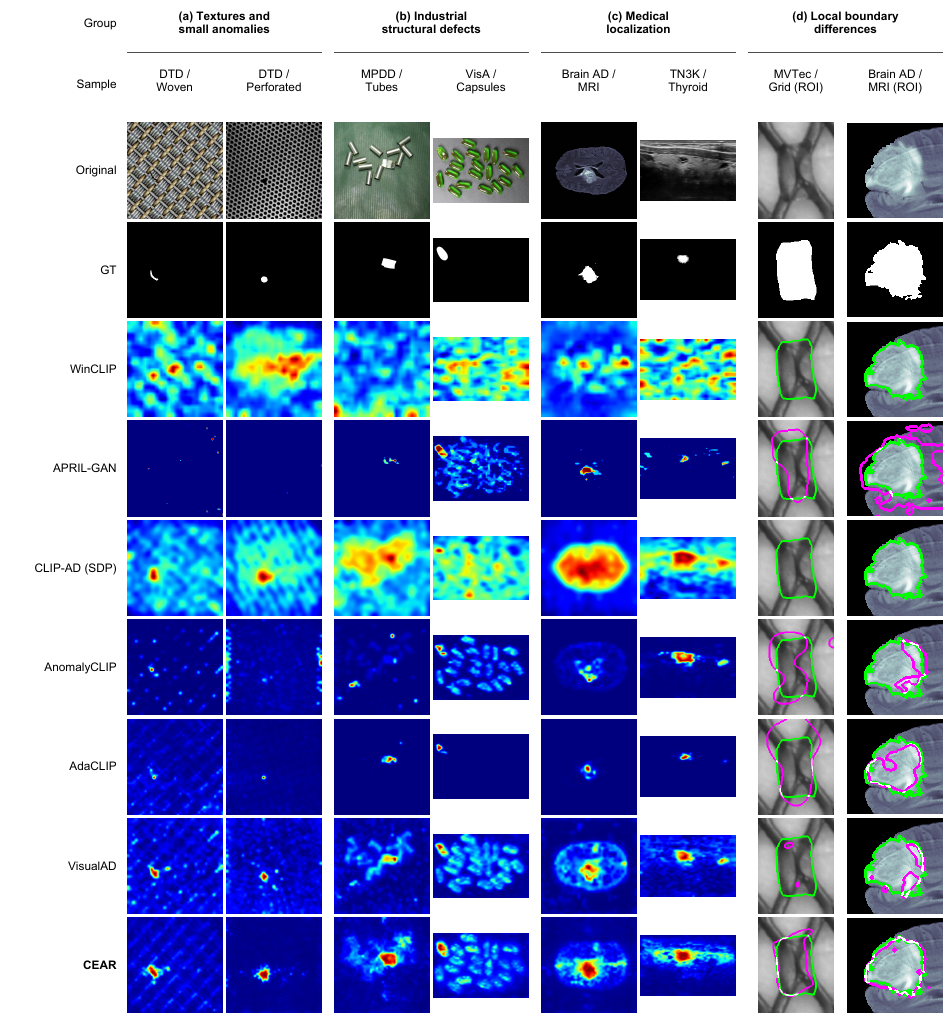}
\caption{Qualitative comparisons for normal textures with small anomalies, industrial structural defects, medical anomaly localization, and local boundary differences. Each group contains two examples. Green and magenta contours denote the ground-truth and predicted boundaries respectively.}\label{fig5}
\end{figure*}

\subsubsection{Refinement Architecture and Interaction Stage}\label{sec4-3-3}

Table~\ref{tab5} removes anchor self-attention, CNN cross-attention, and the feed-forward update individually, and examines parameter sharing across layers and memory compression. Table~\ref{tab6} compares direct use of the refinement output with fixed residual mixing, learned gated mixing, and direct interactions at the patch level.

\begin{table}[!t]
\centering
\caption{Ablation of memory encoding and anchor refinement. SA, CA, and FFN denote self-attention, cross-attention, and feed-forward network, respectively. Boldface and underlining follow the convention used in Table~\ref{tab2}.}
\label{tab5}
\small
\setlength{\tabcolsep}{3pt}
\renewcommand{\arraystretch}{1.15}
\begin{adjustbox}{max width=\columnwidth}
\begin{tabular}{@{} l @{\hspace{10pt}} c @{\hspace{14pt}} c @{}}
\toprule
Configuration & \shortstack{Pixel-level\\(AUROC, F1-max, AP, PRO)} & \shortstack{Image-level\\(AUROC, F1-max, AP)} \\
\midrule
\textbf{Full refinement} & (\makebox[2.3em][r]{\textbf{95.9}}, \makebox[2.3em][r]{\textbf{34.9}}, \makebox[2.3em][r]{\textbf{28.4}}, \makebox[2.3em][r]{\textbf{88.5}}) & (\makebox[2.3em][r]{\textbf{84.8}}, \makebox[2.3em][r]{\textbf{82.7}}, \makebox[2.3em][r]{\textbf{88.5}}) \\
w/o anchor SA & (\makebox[2.3em][r]{95.6}, \makebox[2.3em][r]{34.0}, \makebox[2.3em][r]{26.9}, \makebox[2.3em][r]{87.9}) & (\makebox[2.3em][r]{84.2}, \makebox[2.3em][r]{82.2}, \makebox[2.3em][r]{87.6}) \\
w/o CNN CA & (\makebox[2.3em][r]{95.5}, \makebox[2.3em][r]{33.9}, \makebox[2.3em][r]{\underline{27.3}}, \makebox[2.3em][r]{87.5}) & (\makebox[2.3em][r]{83.7}, \makebox[2.3em][r]{82.0}, \makebox[2.3em][r]{87.7}) \\
w/o FFN & (\makebox[2.3em][r]{95.5}, \makebox[2.3em][r]{34.1}, \makebox[2.3em][r]{26.9}, \makebox[2.3em][r]{\underline{88.3}}) & (\makebox[2.3em][r]{84.2}, \makebox[2.3em][r]{82.3}, \makebox[2.3em][r]{87.9}) \\
Shared refinement & (\makebox[2.3em][r]{95.6}, \makebox[2.3em][r]{\underline{34.5}}, \makebox[2.3em][r]{\underline{27.3}}, \makebox[2.3em][r]{\underline{88.3}}) & (\makebox[2.3em][r]{\underline{84.3}}, \makebox[2.3em][r]{\underline{82.6}}, \makebox[2.3em][r]{\underline{88.0}}) \\
w/o memory SA & (\makebox[2.3em][r]{95.4}, \makebox[2.3em][r]{33.7}, \makebox[2.3em][r]{27.1}, \makebox[2.3em][r]{88.1}) & (\makebox[2.3em][r]{84.2}, \makebox[2.3em][r]{82.2}, \makebox[2.3em][r]{87.7}) \\
Memory 42 \(\times\) 42 & (\makebox[2.3em][r]{\underline{95.7}}, \makebox[2.3em][r]{34.0}, \makebox[2.3em][r]{27.0}, \makebox[2.3em][r]{88.0}) & (\makebox[2.3em][r]{84.1}, \makebox[2.3em][r]{82.2}, \makebox[2.3em][r]{87.7}) \\
Memory 21 \(\times\) 21 & (\makebox[2.3em][r]{95.6}, \makebox[2.3em][r]{33.8}, \makebox[2.3em][r]{26.8}, \makebox[2.3em][r]{87.6}) & (\makebox[2.3em][r]{83.9}, \makebox[2.3em][r]{82.0}, \makebox[2.3em][r]{87.4}) \\
\bottomrule
\end{tabular}
\end{adjustbox}
\end{table}

\begin{table}[!t]
\centering
\caption{Ablation of refinement-output integration and interaction stage. Direct use passes refined anchors to prototype construction while retaining internal residual connections. Patch CA (C4) applies patch-level cross-attention using stride-4 CNN features. Boldface and underlining follow the convention used in Table~\ref{tab2}.}
\label{tab6}
\small
\setlength{\tabcolsep}{3pt}
\renewcommand{\arraystretch}{1.15}
\begin{adjustbox}{max width=\columnwidth}
\begin{tabular}{@{} l @{\hspace{10pt}} c @{\hspace{14pt}} c @{}}
\toprule
Configuration & \shortstack{Pixel-level\\(AUROC, F1-max, AP, PRO)} & \shortstack{Image-level\\(AUROC, F1-max, AP)} \\
\midrule
\textbf{Direct use (default)} & (\makebox[2.3em][r]{\textbf{95.9}}, \makebox[2.3em][r]{\textbf{34.9}}, \makebox[2.3em][r]{\textbf{28.4}}, \makebox[2.3em][r]{\underline{88.5}}) & (\makebox[2.3em][r]{\textbf{84.8}}, \makebox[2.3em][r]{\textbf{82.7}}, \makebox[2.3em][r]{\textbf{88.5}}) \\
Fixed mix (\(\alpha=0.1\)) & (\makebox[2.3em][r]{95.6}, \makebox[2.3em][r]{33.9}, \makebox[2.3em][r]{26.8}, \makebox[2.3em][r]{88.2}) & (\makebox[2.3em][r]{84.0}, \makebox[2.3em][r]{82.3}, \makebox[2.3em][r]{87.5}) \\
Gated mix & (\makebox[2.3em][r]{\underline{95.7}}, \makebox[2.3em][r]{\underline{34.2}}, \makebox[2.3em][r]{\underline{27.1}}, \makebox[2.3em][r]{\underline{88.5}}) & (\makebox[2.3em][r]{84.2}, \makebox[2.3em][r]{82.3}, \makebox[2.3em][r]{87.6}) \\
Memory pre-norm & (\makebox[2.3em][r]{95.6}, \makebox[2.3em][r]{\underline{34.2}}, \makebox[2.3em][r]{\underline{27.1}}, \makebox[2.3em][r]{\textbf{88.6}}) & (\makebox[2.3em][r]{\underline{84.3}}, \makebox[2.3em][r]{\underline{82.4}}, \makebox[2.3em][r]{\underline{87.7}}) \\
Patch CA (C4) & (\makebox[2.3em][r]{95.3}, \makebox[2.3em][r]{34.0}, \makebox[2.3em][r]{27.0}, \makebox[2.3em][r]{88.4}) & (\makebox[2.3em][r]{83.3}, \makebox[2.3em][r]{81.5}, \makebox[2.3em][r]{87.1}) \\
\bottomrule
\end{tabular}
\end{adjustbox}
\end{table}

Removing CNN cross-attention reduces localization performance, while anchor self-attention and the feed-forward update provide further improvements in evidence organization and transformation. Independent refinement parameters at each layer outperform full parameter sharing, supporting depth-specific mappings between ViT representations and local conditioning evidence. Table~\ref{tab6} assesses anchor refinement against direct patch interactions using all four localization metrics and three detection metrics. Fixed residual mixing and learned gated mixing further expose the trade-offs between image-level detection and pixel-level localization. Direct use of the refined anchors yields the highest scores on six metrics, although additional memory pre-normalization gives a slightly higher PRO.

\subsubsection{Supervision across Layers}\label{sec4-3-4}

Table~\ref{tab7} compares individual layers, the three deeper layers, and all four layers. It also removes the query margin, deep region margin, shallow hard-region constraint, and fused-map supervision in turn. Applying the full segmentation loss uniformly to every layer provides a further control for the role of differentiated supervision.

\begin{table}[!t]
\centering
\caption{Ablation of layer selection and supervision. ``All layers'' denotes layers 7, 14, 21, and 27; ``All-layer \(\mathcal{L}_{\mathrm{seg}}\)'' applies segmentation supervision to all four. \(\mathcal{L}_{\mathrm{qry}}\), \(\mathcal{L}_{\mathrm{sem}}\), \(\mathcal{L}_{\mathrm{tex}}\), and \(\mathcal{L}_{\mathrm{map}}\) denote query-separation, deep-region, shallow hard-region, and fused-map losses, respectively. Boldface and underlining follow the convention used in Table~\ref{tab2}.}
\label{tab7}
\small
\setlength{\tabcolsep}{3pt}
\renewcommand{\arraystretch}{1.15}
\begin{adjustbox}{max width=\columnwidth}
\begin{tabular}{@{} l @{\hspace{10pt}} c @{\hspace{14pt}} c @{}}
\toprule
\textbf{Configuration} & \textbf{\shortstack{Pixel-level\\(AUROC, F1-max, AP, PRO)}} & \textbf{\shortstack{Image-level\\(AUROC, F1-max, AP)}} \\
\midrule
\textbf{All layers (default)} & (\makebox[2.3em][r]{\textbf{95.9}}, \makebox[2.3em][r]{\textbf{34.9}}, \makebox[2.3em][r]{\textbf{28.4}}, \makebox[2.3em][r]{\underline{88.5}}) & (\makebox[2.3em][r]{\textbf{84.8}}, \makebox[2.3em][r]{\textbf{82.7}}, \makebox[2.3em][r]{\textbf{88.5}}) \\
Layer 7 only & (\makebox[2.3em][r]{92.0}, \makebox[2.3em][r]{18.9}, \makebox[2.3em][r]{13.5}, \makebox[2.3em][r]{77.1}) & (\makebox[2.3em][r]{67.9}, \makebox[2.3em][r]{76.1}, \makebox[2.3em][r]{73.8}) \\
Layer 27 only & (\makebox[2.3em][r]{93.7}, \makebox[2.3em][r]{27.2}, \makebox[2.3em][r]{19.3}, \makebox[2.3em][r]{83.8}) & (\makebox[2.3em][r]{84.6}, \makebox[2.3em][r]{81.8}, \makebox[2.3em][r]{87.7}) \\
Layers 14/21/27 & (\makebox[2.3em][r]{95.4}, \makebox[2.3em][r]{34.1}, \makebox[2.3em][r]{26.9}, \makebox[2.3em][r]{88.3}) & (\makebox[2.3em][r]{84.5}, \makebox[2.3em][r]{\underline{82.5}}, \makebox[2.3em][r]{87.9}) \\
All-layer \(\mathcal{L}_{\mathrm{seg}}\) & (\makebox[2.3em][r]{95.5}, \makebox[2.3em][r]{34.6}, \makebox[2.3em][r]{27.7}, \makebox[2.3em][r]{88.2}) & (\makebox[2.3em][r]{84.0}, \makebox[2.3em][r]{82.2}, \makebox[2.3em][r]{87.5}) \\
w/o \(\mathcal{L}_{\mathrm{tex}}\) & (\makebox[2.3em][r]{95.5}, \makebox[2.3em][r]{33.3}, \makebox[2.3em][r]{25.4}, \makebox[2.3em][r]{\textbf{88.7}}) & (\makebox[2.3em][r]{\underline{84.7}}, \makebox[2.3em][r]{82.4}, \makebox[2.3em][r]{\underline{88.4}}) \\
w/o \(\mathcal{L}_{\mathrm{sem}}\) & (\makebox[2.3em][r]{\underline{95.6}}, \makebox[2.3em][r]{\underline{34.8}}, \makebox[2.3em][r]{\underline{28.2}}, \makebox[2.3em][r]{88.4}) & (\makebox[2.3em][r]{84.6}, \makebox[2.3em][r]{82.2}, \makebox[2.3em][r]{87.7}) \\
w/o \(\mathcal{L}_{\mathrm{qry}}\) & (\makebox[2.3em][r]{\underline{95.6}}, \makebox[2.3em][r]{34.1}, \makebox[2.3em][r]{26.9}, \makebox[2.3em][r]{88.4}) & (\makebox[2.3em][r]{84.4}, \makebox[2.3em][r]{82.3}, \makebox[2.3em][r]{87.9}) \\
w/o \(\mathcal{L}_{\mathrm{map}}\) & (\makebox[2.3em][r]{95.5}, \makebox[2.3em][r]{33.6}, \makebox[2.3em][r]{26.4}, \makebox[2.3em][r]{88.5}) & (\makebox[2.3em][r]{84.5}, \makebox[2.3em][r]{\underline{82.5}}, \makebox[2.3em][r]{88.2}) \\
\bottomrule
\end{tabular}
\end{adjustbox}
\end{table}

\begin{figure}[!t]
\centering
\includegraphics[width=\columnwidth,height=0.79\textheight,keepaspectratio]{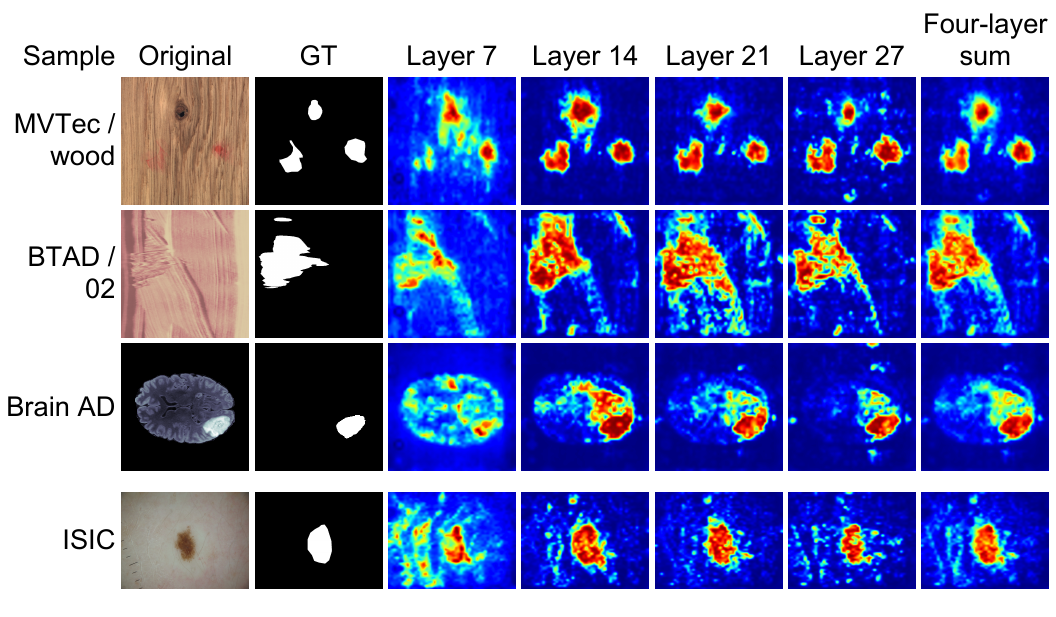}
\caption{Layerwise visualizations examine how responses from different feature depths contribute to CEAR's fused anomaly map. Prediction columns show layers 7, 14, 21, and 27 from the same trained checkpoint, followed by the pointwise sum of their original score maps.}\label{fig6}
\end{figure}

Using the deeper branches preserves overall discrimination, but their small-defect recall remains below that of the full model. The shallow-only branch is more susceptible to responses on normal texture. The shallow hard-region constraint helps suppress local false activations, the deep region margin promotes foreground--background separation, and fused-map supervision supports regional completeness. These contributions do not improve every metric uniformly: removing the shallow region loss slightly increases PRO while reducing the other localization metrics. Taken together, the results support combining complementary layer representations with supervision suited to their respective roles.

\subsection{Qualitative Analysis}\label{sec4-4}

Figure~\ref{fig5} examines localization in four groups of examples: normal textures with small anomalies, industrial structural defects, medical anomalies, and local boundary differences. The comparisons illustrate where CEAR better distinguishes defects from normal appearance and more closely follows annotated regions than the other methods shown.

Figure~\ref{fig6} provides a complementary view of the layerwise behavior discussed in Section~\ref{sec4-3-4}. It compares the branches at layers 7, 14, 21, and 27 with the sum of their scores, all from the same trained checkpoint. Two industrial and two medical examples were selected from existing inference results on domains excluded from anomaly adaptation. They illustrate spatial differences among the individual responses and the fused maps.

\FloatBarrier
\section{Conclusion}\label{sec5}

We presented CEAR, which integrates contextual representations and complementary local evidence through cross-encoder anchor refinement before prototype construction. Spatial anchors read from each selected ViT layer retrieve evidence from a shared local memory built by an independent encoder of the same image. Residual refinement incorporates this evidence before normal and anomalous queries gate anchor aggregation into image-conditioned prototypes. Across the comparisons on 13 industrial and medical datasets, CEAR achieves 57 best results among 68 dataset--metric evaluations, including 41 outright wins in 44 pixel-level comparisons. Across 11 localization datasets, pixel F1-max and AP improve by averages of 4.45 and 3.79 percentage points over the strongest competing result for each dataset. These findings support anchor-stage cross-encoder interaction as a means of integrating complementary visual representations for fine-grained anomaly localization across domains. The shared memory introduces additional computation, motivating future work on memory efficiency and stability under domain shifts.

\FloatBarrier
\bibliographystyle{elsarticle-num-names}
\bibliography{sn-bibliography}

\end{document}